\documentclass{article}

\usepackage{arxiv}

\usepackage[utf8]{inputenc} % allow utf-8 input
\usepackage[T1]{fontenc}    % use 8-bit T1 fonts
\usepackage{amsmath,amssymb,amsfonts}%
\usepackage{amsthm}%
\usepackage{algorithm}%
\usepackage{algorithmicx}%
\usepackage{algpseudocode}%
\usepackage{multirow}
\usepackage{hyperref}       % hyperlinks
\usepackage{cleveref}
\usepackage{url}            % simple URL typesetting
\usepackage{booktabs}       % professional-quality tables
\usepackage{amsfonts}       % blackboard math symbols
\usepackage{nicefrac}       % compact symbols for 1/2, etc.
\usepackage{microtype}      % microtypography
\usepackage{lipsum}		% Can be removed after putting your text content
\usepackage{graphicx}
\usepackage[table]{xcolor}
\usepackage{xcolor}
\usepackage[numbers]{natbib}
\usepackage{doi}

\title{AF-KAN: Activation Function-Based Kolmogorov-Arnold Networks for Efficient Representation Learning}

\author{ \href{https://orcid.org/0000-0003-0321-5106}{\includegraphics[scale=0.06]{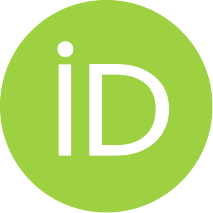}\hspace{1mm}Hoang-Thang Ta}
	\\ Department of Information Technology, Dalat University, Lam Dong, Vietnam \\
	\texttt{thangth@dlu.edu.vn} \\
	   \AND
	\href{https://orcid.org/0000-0002-2903-4373}{\includegraphics[scale=0.06]{orcid.pdf}\hspace{1mm}Anh Tran} \\
	  FPT University, Danang, Vietnam \\
       \texttt{anhtn35@fe.edu.vn} \\ 
}

\renewcommand{\shorttitle}{\textit{arXiv} Template}

\hypersetup{
pdftitle={A template for the arxiv style},
pdfsubject={q-bio.NC, q-bio.QM},
pdfauthor={David S.~Hippocampus, Elias D.~Striatum},
pdfkeywords={First keyword, Second keyword, More},
}

\begin{document}
\maketitle

\begin{abstract}
Kolmogorov-Arnold Networks (KANs) have inspired numerous works exploring their applications across a wide range of scientific problems, with the potential to replace Multilayer Perceptrons (MLPs). While many KANs are designed using basis and polynomial functions, such as B-splines, ReLU-KAN utilizes a combination of ReLU functions to mimic the structure of B-splines and take advantage of ReLU's speed. However, ReLU-KAN is not built for multiple inputs, and its limitations stem from ReLU’s handling of negative values, which can restrict feature extraction. To address these issues, we introduce Activation Function-Based Kolmogorov-Arnold Networks (AF-KAN), expanding ReLU-KAN with various activations and their function combinations. This novel KAN also incorporates parameter reduction methods, primarily attention mechanisms and data normalization, to enhance performance on image classification datasets. We explore different activation functions, function combinations, grid sizes, and spline orders to validate the effectiveness of AF-KAN and determine its optimal configuration. In the experiments, AF-KAN significantly outperforms MLP, ReLU-KAN, and other KANs with the same parameter count. It also remains competitive even when using fewer than 6 to 10 times the parameters while maintaining the same network structure. However, AF-KAN requires a longer training time and consumes more FLOPs. The repository for this work is available at \url{https://github.com/hoangthangta/All-KAN}.

\end{abstract}
%(PRKAN-CNN) 
% keywords can be removed

\section{Introduction}
Recently, Kolmogorov-Arnold Networks (KANs) have gained significant attention from the research community due to their innovative approach to data representation in neural networks~\cite{liu2024kan, liu2024kan2.0}. Unlike traditional MLPs, which use fixed activation functions as "nodes", KANs apply learnable functions to "edges". This characteristic has inspired researchers to explore novel network architectures to evaluate KANs' effectiveness across various problems. The motivation behind this research stems not only from scientific curiosity but also from the long-standing dominance of MLPs in neural networks. It is simply time to rethink the role of "old" MLPs. The principle behind KANs is the Kolmogorov-Arnold Representation Theorem~\cite{kolmogorov1957representation}, formulated to address Hilbert's 13th problem~\cite{sternfeld2006hilbert}. This theorem states that any multivariate function can be expressed as a sum of continuous single-variable functions, forming the theoretical foundation for KANs' function decomposition, which enhances flexibility and efficiency in certain applications.

From the use of B-splines in the original KAN~\cite{liu2024kan}, numerous studies have explored replacing them with various polynomial and basis functions~\cite{li2024kolmogorov,abueidda2024deepokan,ta2024bsrbf,torchkan,ss2024chebyshev,xu2024fourierkan,bozorgasl2024wav,seydi2024unveiling,aghaei2024rkan,aghaei2024fkan,teymoor2024exploring} to enhance their understanding of these functions in various problems. B-splines are not fully optimized for GPU acceleration, and networks utilizing them can be slower than activation functions used in traditional MLPs. Researchers have revisited standard activation functions such as ReLU and Tanh to address this limitation, integrating them into KAN-style architectures through function combinations~\cite{qiu2024relu, athanasios2024}. Besides, trigonometric functions have been introduced (LArctan-SKAN~\cite{chen2024larctan}, LSS-SKAN~\cite{chen2024lss}) to reduce both training time and parameter count.

KANs not only exhibit slower training speeds but also suffer from an excessive number of parameters compared to MLPs when using the same network structure. This is due to the nature of curve-based functions in KANs, which require more control points to capture direction and represent data features accurately. \citet{yu2024kan} concluded that while KANs excel in symbolic representation, MLPs outperform them in many other tasks. Given the same network structure, it is unsurprising that KANs achieve better performance than MLPs, as they inherently utilize a significantly larger number of parameters~\cite{ta2024bsrbf, ta2024fc, yang2024activation, moradi2024kolmogorov, sohail2024training}. However, even when KANs are designed with fewer parameters than MLPs, they still require longer training times~\cite{shuai2024physics}.

In this paper, we propose AF-KAN (Activation Function-Based Kolmogorov-Arnold Networks), which builds upon ReLU-KAN—a model that leverages ReLU combinations (such as the square of ReLU) and matrix operations, to enhance representation learning across various image-classification datasets. Unlike ReLU-KAN, AF-KAN incorporates a broader range of widely used activation function combinations beyond ReLU and integrates attention mechanisms to reduce the number of parameters, making its layers comparable to those of MLPs. Instead of normalizing data using a function’s maximum value, we employ an approach that combines the L2 norm and min-max scaling at the batch level, eliminating the need to compute the maximum across different function types. Moreover, we introduce pre-linear normalization before the linear transformation to enhance model performance. In the experiments, we compare AF-KAN against MLPs and other KANs with the similar parameter count and/or the same network structure. Moreover, we evaluate AF-KAN with various activation functions, function types, grid sizes, and spline orders to search for optimal configuration.

In summary, our main contributions are:
\begin{itemize}
    \item Design AF-KAN which is based on ReLU-KAN with  attention mechanisms and data normalization, aiming to reduce the number of parameters and maintain a controlled data value range, all while preserving an identical network structure compared to MLP layers.
    \item Evaluate the effectiveness of AF-KANs by contrasting their performance with MLPs and other KANs, using models trained on the MNIST and Fashion-MNIST image datasets.
    \item Compare model performance across function types, activation functions, and normalization to gain deeper insights into AF-KAN and identify its optimal configuration.
\end{itemize}

Apart from this section, the paper is structured as follows: Section 2 reviews related works on KART and KANs, and the efforts to improve KANs. Section 3 presents our methodology, including details on KART, KANs, parameter count in KANs and MLPs, ReLU-KANs, and AF-KANs. Section 4 presents our experimental results, comparing AF-KANs with MLPs and other KANs on the MNIST and Fashion-MNIST datasets by the same parameter count and/or the same network structure. This section also includes ablation studies on activation functions, function types, grid sizes, spline orders, and data normalization used in AF-KANs. Section 5 discusses some limitations of this study, while Section 6 concludes the paper and suggests possible directions for future research.

\section{Related Works}\label{sec_related_works}

\subsection{KART and KAN}

In 1957, Kolmogorov resolved Hilbert’s 13th problem by demonstrating that any multivariate continuous function can be expressed as a composition of single-variable functions and summations, a principle known as the Kolmogorov–Arnold Representation Theorem (KART)~\cite{kolmogorov1957representation,braun2009constructive}. This theorem has played a significant role in advancing neural networks~\cite{zhou2022treedrnet,leni2013kolmogorov,lai2021kolmogorov,van2022kasam}. Despite its long-standing application in neural networks, KART remained relatively unnoticed in the research community until the recent contributions of \citet{liu2024kan,liu2024kan2.0}. They proposed an extension beyond the conventional KART framework by introducing KANs, which incorporate additional neurons and layers. This perspective aligns with our thought as it effectively mitigates the challenges posed by non-smooth functions in neural networks leveraging KART. Consequently, KANs have the potential to outperform MLPs in both accuracy and interpretability, particularly in small-scale AI + Science applications.

By introducing a novel perspective on neural network architecture, KANs have proven effective in a wide range of studies, addressing various challenges such as computational efficiency~\cite{hao2024first}, solving differential equations~\cite{wang2024kolmogorov, koenig2024kan}, keyword spotting~\cite{xu2024effective}, mechanics-related problems~\cite{abueidda2024deepokan}, quantum computing applications~\cite{kundu2024kanqas, wakaura2024variational, troy2024sparks}, survival analysis~\cite{knottenbelt2024coxkan}, time series forecasting~\cite{genet2024tkan, xu2024kolmogorov, vaca2024kolmogorov, genet2024temporal, han2024kan4tsf}, and vision-related tasks~\cite{li2024u, cheon2024demonstrating, ge2024tc}. These contributions highlight the versatility and robustness of KANs in addressing complex real-world problems across diverse scientific and engineering disciplines.

Various basis and polynomial functions have been utilized in recent KANs~\cite{somvanshi2024survey}, particularly those well-suited for curve representation, such as B-splines~\cite{de1972calculating} (Original KAN~\cite{liu2024kan}, EfficientKAN~\cite{Blealtan2024}, BSRBF-KAN~\cite{ta2024bsrbf}), Gaussian Radial Basis Functions (GRBFs) (FastKAN~\cite{li2024kolmogorov}, DeepOKAN~\cite{abueidda2024deepokan}, BSRBF-KAN~\cite{ta2024bsrbf}), Chebyshev polynomials (TorchKAN~\cite{torchkan}, Chebyshev KAN~\cite{ss2024chebyshev}), Legendre polynomials (TorchKAN~\cite{torchkan}), Fourier transforms (FourierKAN\footnote{https://github.com/GistNoesis/FourierKAN/}, FourierKAN-GCF~\cite{xu2024fourierkan}), wavelets~\cite{bozorgasl2024wav,seydi2024unveiling}, rational functions~\cite{aghaei2024rkan}, fractional Jacobi functions~\cite{aghaei2024fkan}, and other polynomial functions~\cite{teymoor2024exploring}. Additionally, several studies have employed trigonometric functions (LArctan-SKAN~\cite{chen2024larctan}, LSS-SKAN~\cite{chen2024lss}) and custom activation functions (ReLU-KAN~\cite{qiu2024relu}, Reflection Switch Activation Function (RSWAF) in FasterKAN~\cite{athanasios2024}) in the design of KAN architectures.

KANs have demonstrated their versatility by being integrated into various neural network architectures, including autoencoders~\cite{moradi2024kolmogorov}, GNNs~\cite{bresson2024kagnns, de2024kolmogorov, zhang2024graphkan}, Reinforcement Learning~\cite{kich2024kolmogorov}, Transformers~\cite{yang2024kolmogorov}, CNNs, and RNNs. In CNNs, KANs can replace convolutional layers, MLP layers, or a combination of both, offering flexible design possibilities~\cite{abd2024ckan, bodner2024convolutional}. In RNNs, KAN layers are not standalone components; instead, they are integrated with linear weights, fully connected layers, and additional elements such as biases and previous hidden states to form complete architectures~\cite{genet2024tkan, genet2024temporal, danish2025kolmogorov}.

\subsection{Efforts to Improve KANs}

While KANs have been proven effective in a wide range of problems, they still have some disadvantages, which can be classified into two main issues: long training time and parameter inefficiency. For the first issue, the original KAN and its variants utilize polynomial and basis functions such as B-splines, which are not fully supported by GPU devices \cite{liu2024kan}, resulting in significantly slower training speeds compared to activation functions in MLPs. \citet{qiu2024relu,so2024higher} introduce the use of traditional activation functions like ReLU replaced for B-Splines in their KAN networks (ReLU-KAN and HRKAN) to accelerate the training speed. 
%RBFs, and wavelets,

Regarding the second issue, KANs indeed require significantly more parameters than MLPs, which may naturally contribute to their higher performance. KANs have demonstrated efficient parameter utilization in specific applications, such as satellite traffic forecasting \cite{vaca2024kolmogorov} and quantum architecture search \cite{kundu2024kanqas}, where they outperform MLPs without requiring parameter reduction. However, in tasks with comparable parameter budgets and computational complexity, MLPs generally achieve better performance, except in symbolic formula representation tasks \cite{yu2024kan}. From a software and hardware implementation perspective, MLPs remain a more practical choice for achieving high accuracy, as KANs struggle with highly complex datasets while consuming significantly more hardware resources \cite{le2024exploring}.

Parameter reduction has become essential in realizing the full potential of Kolmogorov-Arnold Networks (KANs). \citet{bodner2024convolutional} introduced Convolutional Kolmogorov-Arnold Networks (Convolutional KANs), integrating learnable non-linear activation functions into convolutions, achieving accuracy similar to CNNs with half the parameters, thus improving learning efficiency. CapsuleKAN enhances precision and parameter efficiency in traditional capsule networks by employing ConvKAN and LinearKAN \cite{mou2024efficient}. ConvKAN applies B-spline convolutions to improve feature extraction, while LinearKAN uses B-splines as activation functions to capture non-linearities with fewer parameters. More recently, \citet{ta2025prkan} introduced PRKAN, a novel network that applies various parameter reduction methods, achieving a parameter count comparable to that of Multi-Layer Perceptrons (MLPs).

Especially, several studies have simultaneously addressed both challenges associated with KANs. Two studies by the same authors introduced Single-Parameterized Kolmogorov-Arnold Networks (SKANs), which incorporate basis functions with a single learnable parameter \cite{chen2024lss, chen2024larctan}. These works proposed multiple SKAN variants, including LSS-SKAN, LSin-SKAN, LCos-SKAN, and LArctan-SKAN, demonstrating significant improvements in parameter efficiency and computational performance. In the work of designing Kolmogorov-Arnold Transformers (KATs), \citet{yang2024kolmogorov} introduced Group KAN, a KAN designed to reduce parameter count and computational cost by employing shared parameters within groups of edges for each input-output pair.

%We also identified studies that, while not directly focused on parameter reduction in KANs, provide valuable insights related to this topic. For instance, rather than focusing on KANs—which require significantly more parameters and perform less effectively in data-scarce domains—\citet{pourkamali2024kolmogorov} explore MLPs with parameterized activation functions for each neuron. Their experiments show that MLPs achieve higher accuracy than KANs with only a modest increase in parameters. To improve training speed, \citet{qiu2024relu} introduced ReLU-KAN, replacing B-splines with matrix addition, dot products, and ReLU activation to enable efficient GPU parallelization. It also uses exactly two parameters traditionally associated with KANs: grid size and spline order. This method accelerates backpropagation, enhances accuracy, and improves resilience to catastrophic forgetting, all while keeping parameter settings constant. \citet{zinage2024dkl} integrated deep kernel learning (DKL) into KANs (DKL-KAN) and MLPs (DKL-MLP), designing DKL-KAN variants: one with the same neurons and layers as DKL-MLP, and another with a similar number of trainable parameters. They found that DKL-KAN performs better on small datasets, especially in modeling discontinuities and estimating uncertainty, while DKL-MLP excels in scalability and accuracy on larger datasets.

\section{Methodology}
\label{sec:methodology}

\subsection{Kolmogorov-Arnold Network}
\label{KAN_design}

\subsubsection{Kolmogorov-Arnold Representation Theorem}

A KAN is built upon KART, which asserts that any continuous multivariate function $f$, defined over a bounded domain, can be decomposed into a finite sum of continuous single-variable functions~\cite{chernov2020gaussian,schmidt2021kolmogorov}. Given a set of variables $\mathbf{x} = \{x_1, x_2, \ldots, x_n\}$, where $n$ denotes the number of variables, the function $f(\mathbf{x})$ can be formulated as:

\begin{equation}
\begin{aligned}
f(\mathbf{x}) = f(x_1, \ldots, x_n) = \sum_{q=1}^{2n+1} \Phi_q \left( \sum_{p=1}^{n} \phi_{q,p}(x_p) \right) 
\end{aligned}
\label{eq:kart}
\end{equation}

A continuous multivariate function \( f(\mathbf{x}) \) can be represented as a sum of outer functions \( \Phi_q \), each applied to an inner summation of transformed input variables \( x_p \) through functions \( \phi_{q,p} \). This formulation consists of two levels of summation: the outer sum, \( \sum_{q=1}^{2n+1} \), which combines \( 2n+1 \) continuous functions \( \Phi_q \) (\(\mathbb{R} \to \mathbb{R}\)), and the inner sum, which aggregates \( n \) terms for each \( q \), where each term \( \phi_{q,p} \) (\(\phi_{q,p} \colon [0,1] \to \mathbb{R}\)) represents a continuous transformation of a single variable \( x_p \). This decomposition enables the representation of any continuous multivariate function using only single-variable functions and summations, forming the theoretical foundation of Kolmogorov-Arnold Networks (KANs).

\subsubsection{Formation and Structure of KAN}
An MLP consists of a sequence of affine transformations followed by nonlinear activation functions. Given an input \( \mathbf{x} \), the network processes it through multiple layers, where a weight matrix and a bias vector define each layer. For a network with \( L \) layers (indexed from \( 0 \) to \( L-1 \)), the transformation at layer \( l \) is expressed as:

\begin{equation}
\begin{aligned}
\text{MLP}(\mathbf{x}) &= (W_{L-1} \circ \sigma \circ W_{L-2} \circ \sigma \circ \cdots \circ W_1 \circ \sigma \circ W_0) \mathbf{x} \\
\end{aligned}
\label{eq:mlp}
\end{equation}

\citet{liu2024kan} developed the Kolmogorov-Arnold Network (KAN) and recommended increasing both its width and depth to enhance its expressive power. This approach relies on the careful selection of functions \( \Phi_q \) and \( \phi_{q,p} \), as defined in \Cref{eq:kart}. In a typical KAN with \( L \) layers, the input \( \mathbf{x} \) undergoes successive transformations through function matrices \( \Phi_0, \Phi_1, \dots, \Phi_{L-1} \), ultimately producing the final output \( \text{KAN}(\mathbf{x}) \), as given by:

\begin{equation}
\begin{aligned}
\text{KAN}(\mathbf{x}) = (\Phi_{L-1} \circ \Phi_{L-2} \circ \cdots \circ \Phi_1 \circ \Phi_0)\mathbf{x}
\end{aligned}
\label{eq:kan}
\end{equation}
The function matrix \( \Phi_l \) at the \( l^{th} \) KAN layer consists of a set of pre-activations. Consider the \( i^{th} \) neuron in the \( l^{th} \) layer and the \( j^{th} \) neuron in the \( (l+1)^{th} \) layer. The activation function \( \phi_{l,i,j} \) defines the connection between neuron \( (l, i) \) and neuron \( (l+1, j) \), expressed as:

%Say the neuron $i^{th}$ of the layer $l^{th}$ and the neuron $j^{th}$ of the layer $l+1^{th}$. The activation function $\phi_{l,i,j}$ connects $(l, i)$ to $(l + 1, j)$:

\begin{equation}
\begin{aligned}
\phi_{l,j,i}, \quad l = 0, \cdots, L - 1, \quad i = 1, \cdots, n_l, \quad j = 1, \cdots, n_{l+1}
\end{aligned}
\label{eq:acti_funct}
\end{equation}
Let \( n_l \) denote the number of nodes in the \( l^{th} \) layer. The input \( \mathbf{x}_l \) is processed through the function matrix \( \Phi_l \), which has dimensions \( n_{l+1} \times n_l \), to compute the output \( \mathbf{x}_{l+1} \) at the \( (l+1)^{th} \) layer, as given by:

\begin{equation}
\begin{aligned}
\mathbf{x}_{l+1} = 
\underbrace{\left(
\begin{array}{cccc}
\phi_{l,1,1}(\cdot) & \phi_{l,1,2}(\cdot) & \cdots & \phi_{l,1,n_l}(\cdot) \\
\phi_{l,2,1}(\cdot) & \phi_{l,2,2}(\cdot) & \cdots & \phi_{l,2,n_l}(\cdot) \\
\vdots & \vdots & \ddots & \vdots \\
\phi_{l,n_{l+1},1}(\cdot) & \phi_{l,n_{l+1},2}(\cdot) & \cdots & \phi_{l,n_{l+1},n_l}(\cdot)
\end{array}\right)}_{\Phi_{l}} \mathbf{x}_l
\label{eq:function_matrix}
\end{aligned}
\end{equation}

%https://arxiv.org/html/2408.07314v1/extracted/5790843/images/mainpic.png
%https://arxiv.org/html/2404.19756v3/extracted/5619424/figs/spline_notation.png

\begin{figure*}[htbp]
  \centering
\includegraphics[scale=0.95]{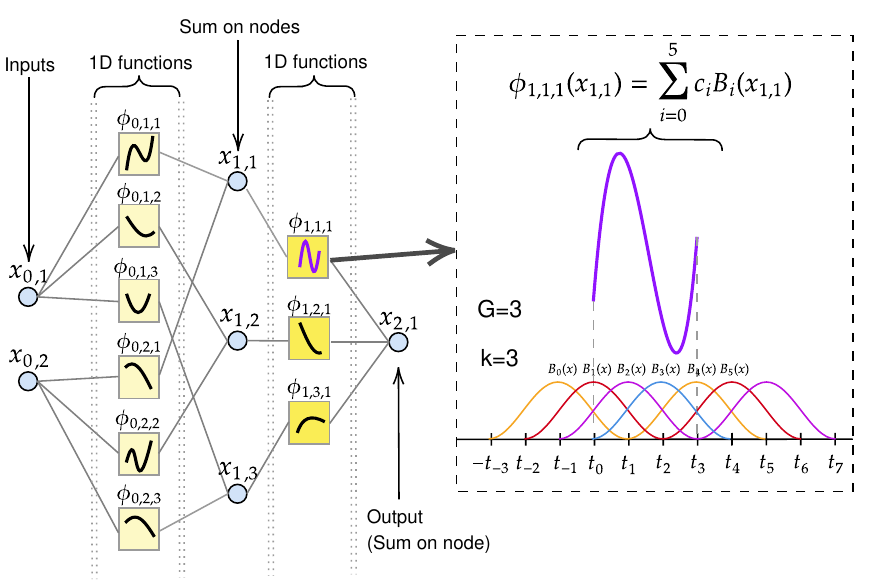}
  \centering
  \caption{Left: The structure of KAN(2,3,1). Right: Calculate \(\phi_{1,1,1}\) by using control points and B-splines~\cite{ta2024fc}. \(G \) and \( k \) represent the grid size and the spline order, while the number of B-splines, \(n\), is given by \( G + k = 3 + 3 = 6\).}
\label{fig:kan_diagram}
\end{figure*}

\subsubsection{Implementation of Existing KANs}
%LiuKAN\footnote{We refer to the original KAN as LiuKAN, named after the first author's last name~\cite{liu2024kan}, while another study~\cite{bozorgasl2024wav} calls it Spl-KAN.}, as implemented by \citet{liu2024kan}, 
~\citet{liu2024kan} constructed KAN by utilizing a residual activation function \(\phi(x)\), which consists of the summation of a base function and a spline function, each associated with weight matrices \(w_b\) and \(w_s\), respectively.

\begin{equation}
\begin{aligned}
\phi(x) = w_b b(x) + w_s spline(x)
\end{aligned}
\label{eq:acti_funct_imp}
\end{equation}

\begin{equation}
\begin{aligned}
b(x) = silu(x) = \frac{x}{1 + e^{-x}}
\end{aligned}
\label{eq:b_function}
\end{equation}

\begin{equation}
\begin{aligned}
spline(x) = \sum_{i}c_iB_i(x)
\end{aligned}
\label{eq:spline_function}
\end{equation}
In \Cref{eq:acti_funct_imp}, \( b(x) \) equals \( silu(x) \) (as presented in \Cref{eq:b_function}), whereas \( spline(x) \) is expressed as a linear combination of B-splines \( B_i \) and their corresponding control points or coefficients \( c_i \) (as illustrated in \Cref{eq:spline_function}). Then, the activation functions become active by setting \( w_s = 1 \), which keeps \( spline(x) \approx 0 \), while \( w_b \) is initialized using the Xavier initialization. It is worth mentioning that alternative initializations, such as Kaiming may also be considered~\cite{yang2024kolmogorov}.

\Cref{fig:kan_diagram} depicts the architecture of KAN(2,3,1), which includes 2 input nodes, 3 hidden nodes, and 1 output node. Each node’s output is computed as the summation of individual functions \(\phi\), represented as "edges". The figure also illustrates the process of computing the inner function \(\phi\) by control points and B-splines. The number of B-splines is determined by summing the grid size \(G\) and the spline order \(k\), yielding \( G + k = 3 + 3 = 6 \), meaning the index \(i\) ranges from 0 to 5.

\textbf{EfficientKAN} adopts a similar methodology to \citet{liu2024kan}, but it optimizes computations by leveraging B-splines and linear combinations, thereby reducing memory consumption and simplifying calculations~\cite{Blealtan2024}. The previous L1 regularization applied to input samples was replaced with L1 regularization on the weights. Additionally, learnable scaling factors for the activation functions were incorporated, and the initialization of both the base weight and spline scaling matrices was modified to Kaiming uniform initialization.

\textbf{FastKAN} accelerates training relative to EfficientKAN by employing RBFs to approximate the third-order B-spline and integrating layer normalization to keep inputs within the RBFs' domain~\cite{li2024kolmogorov}. These enhancements streamline the implementation while preserving accuracy. The RBF is defined as follows:

\begin{equation}
\begin{aligned}
\phi(r) = e^{-\epsilon r^2}
\end{aligned}
\label{eq:gaussian_rbf}
\end{equation}

The distance between an input vector \( x \) and a center \( c \) is represented as \( r = \|x - c\| \), where \( \epsilon \) (\( \epsilon > 0 \)) is a parameter that controls the width of the Gaussian function. In FastKAN, Gaussian Radial Basis Functions (GRBFs) are employed, with \( \epsilon \) set to \( \frac{1}{2h^2} \), as explained in~\cite{li2024kolmogorov}, and defined by:

\begin{equation}
\begin{aligned}
\phi_{\mathit{RBF}}(r) = \exp\left(-\frac{r^2}{2h^2}\right)
\end{aligned}
\label{eq:special_gaussian_rbf}
\end{equation}
The parameter \( h \) determines the width of the Gaussian function. Consequently, the RBF network with \( C \) centers can be formulated as~\cite{li2024kolmogorov,ta2024bsrbf,ta2024fc}:

\begin{equation}
\begin{aligned}
RBF(x) = \sum_{i=1}^{C} w_i \phi_{\mathit{RBF}}(r_i) = \sum_{i=1}^{C} w_i \exp\left(-\frac{||x - c_i||}{2h^2}\right)
\end{aligned}
\label{eq:rbf_network}
\end{equation}
The weight \( w_i \) represents the trainable coefficients, while \( \phi \) denotes the radial basis function, as outlined in \Cref{eq:gaussian_rbf}.

Compared to FastKAN, \textbf{FasterKAN} shows improved processing speeds in both forward and backward passes~\cite{athanasios2024}. It utilizes Reflectional Switch Activation Functions (RSWAFs), which are streamlined versions of RBFs and are computationally efficient due to their uniform grid structure. The RSWAF function is defined as:

\begin{equation}
\begin{aligned}
\phi_{\mathit{RSWAF}}(r) = 1 - \left(\tanh\left(\frac{r}{h}\right)\right)^2
\end{aligned}
\label{eq:rswaf_funct}
\end{equation}

The network with $N$ centers is expressed as:
\begin{equation}
\begin{aligned}
\mathit{RSWAF}(x) = \sum_{i=1}^{N} w_i \phi_{\mathit{RSWAF}}(r_i) = \sum_{i=1}^{N} w_i \left(1 - \left(\tanh\left(\frac{||x - c_i||}{h}\right)\right)^2\right)
\end{aligned}
\label{eq:rswaf_network}
\end{equation}

\textbf{BSRBF-KAN} is a KAN variant incorporating B-splines from EfficientKAN and GRBFs from FastKAN in each layer through additive operations. It achieves faster convergence during training than EfficientKAN, FastKAN, and FasterKAN. However, this characteristic may result in overfitting and does not necessarily ensure high validation accuracy~\cite{ta2024bsrbf}. The BSRBF function is formulated as:

\begin{equation}
\begin{aligned}
\phi_{BSRBF}(x)  =  w_b b(x) + w_s (\phi_{BS}(x) + \phi_{RBF}(x)) 
\end{aligned}
\label{eq:bsrbf_funct}
\end{equation}
The base function \( b(x) \) and its corresponding matrix \( w_b \) represent the linear component present in traditional MLP layers, while \( \phi_{BS}(x) \) and \( \phi_{RBF}(x) \) relate to the B-spline and Radial Basis Function (RBF), respectively. The matrix \( w_s \) is linked to the coefficients associated with the sum of \( \phi_{BS}(x) \) and \( \phi_{RBF}(x) \).

\textbf{FC-KAN} is a Kolmogorov-Arnold Network (KAN) that utilizes combinations of popular mathematical functions, such as B-splines, wavelets, and radial basis functions on low-dimensional data through element-wise operations~\cite{ta2024fc}. It employs various methods to combine the outputs of these functions, including summation, element-wise product, the addition of sum and element-wise product, representations of quadratic and cubic functions, concatenation, linear transformation of the concatenated output, and additional approaches.

\subsubsection{Parameter Requirements in KANs vs. MLPs}

In this section, we present the parameter gap between KANs and MLPs to explain why KANs use significantly more parameters than MLPs. To ensure a fair comparison and explore their capabilities, it is necessary to reduce these parameters in KANs. As mentioned in~\citet{ta2025prkan}, consider an input \( x \) and a network layer with an input dimension of \( d_{\text{in}} \) and an output dimension of \( d_{\text{out}} \). Let \( k \) denote the spline order and \( G \) represent the grid size of a function, such as B-splines, used in Kolmogorov-Arnold Networks (KANs).

The required number of control points, which also corresponds to the number of basis functions, is \( G + k \). The total number of parameters, encompassing the weight matrix and the bias term, when processing \( x \) through a KAN layer, is:

\begin{equation}
\begin{aligned}
KAN_{\text{params}} = 
\underbrace{d_{in} \times d_{out} \times (G + k)}_{\text{weight matrix params}} 
+ 
\underbrace{d_{out}}_{\text{bias matrix params}}
\end{aligned}
\label{eq:kan_params}
\end{equation}
while an MLP layer requires:

\begin{equation}
\begin{aligned}
MLP_{\text{params}} = 
\underbrace{d_{in} \times d_{out}}_{\text{weight matrix params}} 
+ 
\underbrace{d_{out}}_{\text{bias matrix params}}
\end{aligned}
\label{eq:mlp_params}
\end{equation}
Note that we omit other additional parameters used in KANs and only retain the important ones. As a result, the number of parameters is lower than that in the work of~\citet{yu2024kan}. As shown in \Cref{eq:kan_params} and \Cref{eq:mlp_params}, KANs consistently require more parameters than MLPs, making direct comparisons in networks with the same layer structure inequitable.

%For example, given a list with 4 data points, \([0.4, 0.5, 0.6, 0.7]\), when passing this list to the Sigmoid function ($\sigma(x) = 1/(1 + e^{-x})$), the output is \(\textit{tensor}\left( \begin{bmatrix} \begin{bmatrix} 0.5986, 0.6224, 0.6456, 0.6681 \end{bmatrix} \end{bmatrix} \right)\), with shape \((1, 4)\). When doing this with a B-spline in EfficientKAN\footnote{The output has the same features (shape and partition of unity) when using Liu-KAN.}, the output is as follows:

We reuse an example from ~\cite{ta2025prkan}. Consider a list of 4 data points, \([0.4, 0.5, 0.6, 0.7]\). When applying the Sigmoid function (\(\sigma(x) = 1/(1 + e^{-x})\)) to this list, the resulting output is \(\textit{tensor} \left( \begin{bmatrix} 0.5986, 0.6224, 0.6456, 0.6681 \end{bmatrix} \right)\), with a shape of \((1,4)\). Similarly, when processing this list using a B-spline in EfficientKAN\footnote{The output retains the same properties (shape and partition of unity) when using the original KAN~\cite{liu2024kan}}, the resulting output is:

\[
\textit{tensor}\left(
\begin{bmatrix}
\begin{bmatrix}
\begin{bmatrix}0.0000, 0.0000, 0.0000, 0.0208, 0.4792, 0.4792, 0.0208, 0.0000\end{bmatrix}, \\
\begin{bmatrix}0.0000, 0.0000, 0.0000, 0.0026, 0.3151, 0.6120, 0.0703, 0.0000 \end{bmatrix},\\
\begin{bmatrix}0.0000, 0.0000, 0.0000, 0.0000, 0.1667, 0.6667, 0.1667, 0.0000 \end{bmatrix},\\
\begin{bmatrix}0.0000, 0.0000, 0.0000, 0.0000, 0.0703, 0.6120, 0.3151, 0.0026 \end{bmatrix} \

\end{bmatrix}
\end{bmatrix}
\right)
\]
This tensor has a $(1, 4, 8)$ shape, where the batch size is 1, 4 data points are processed, and 8 basis functions  ($G + k$ = 5 + 3) are evaluated per input. Each row sums approximately to 1 due to the partition of unity property of B-spline basis functions, which ensures smoothness, locality, and a complete representation of the input. KANs naturally use more parameters than MLPs, enabling them to capture data features effectively.

\subsection{ReLU-KAN}
ReLU-KAN uses a set of "ReLU combination" functions  
\( \textbf{R} = \{R_1(x), R_2(x), \dots, R_n(x)\} \), which follows the style of the original KAN~\cite{liu2024kan} with B-splines. It includes \( n = G + k\) basis functions of the same bell shape at different locations, along with the grid size \( G \) and function degree \( k \). Each function $R_i(x)$ has the formula~\cite{qiu2024relu}:

\begin{equation}
\begin{aligned}
R_i(x) = [ReLU(x - l_i) \times ReLU(h_i - x)]^2 \times 16/(h_i - l_i)^4
\end{aligned}
\label{eq:relu_kan}
\end{equation}
which $ReLU(x) = max(0, x)$ and the function is nonzero at $x \in
[l_i, h_i]$ and zero at other ranges. It also uses phase high $l_i$ and phase low $h_i$ are trainable parameters. The initial values of \( l_i \) and \( h_i \) are given by \( l_i = \frac{-k + i - 1}{G} \) and \( h_i = \frac{i}{G} \), respectively. With the plot of bell-shape, $R_i(x)$ has the maximum value $m = (h_i - l_i)^4/16$, and the constant value  $c = 16/(h_i - l_i)^4$ is used to normalize the value range. 

\begin{figure*}[htbp]
  \centering
\includegraphics[scale=0.9]{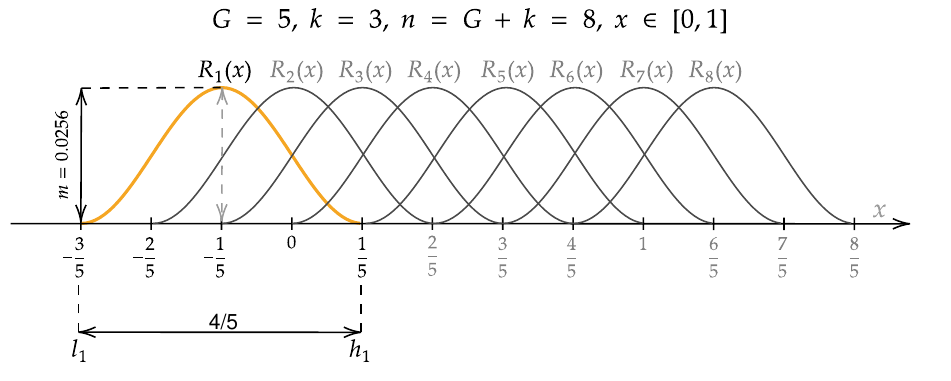}
  \centering
  \caption{Simulate the plot of $\textbf{R}$ with a grid size $G = 5$ and a spline order $k = 3$.}
\label{fig:relu_kan}
\end{figure*}

For example, if \( G = 5 \) and \( k = 3 \), the function \( \textbf{R} \) has 8 spline functions (\( n = G + k = 5 + 3 = 8\)) with the domain \( x \in [0,1] \). For \( R_1(x) \), we have the interval of the bell-shaped part from $l_1 = -3/5$ to $h_1 = 1/5$, the maximum value $m = (3/5 + 1/5)^4/16 = 0.0256$, and the normalization constant $c = 1/m = 39.0625$. Note that $m$ and $c$ are the same for other $R_i(x)$ functions. The demo for this example is shown in \Cref{fig:relu_kan}.

The original ReLU-KAN could only process single inputs~\cite{qiu2024relu}. Therefore, we modified it to handle multiple inputs by expanding the input to match the sizes of phase low and phase high, which enables matrix operations. This modified ReLU-KAN is, of course, utilized for the experiments in this paper.

%We broaden the use of additional activation functions, not just ReLU; however, ReLU remains the default setting. 
%We also added two types of data normalization, layer and batch, and broadened the use of other activation functions, not just ReLU. However, no normalization and ReLU remain the default settings. And, of course, this modified ReLU-KAN is used for the experiments in this paper.

\subsection{AF-KAN}
AF-KAN (\textbf{A}ctivation \textbf{F}unction-Based \textbf{K}olmogorov-\textbf{A}rnold \textbf{N}etworks) is derived from ReLU-KAN but is generalized to support various activation functions (ReLU, SiLU, GeLU, ELU, SeLU, etc.) and their combination types, extending from single functions to cubic forms. This network also applies attention mechanisms and data normalization to reduce the number of parameters and improve model performance.

Instead of B-Splines, we use a set of functions $\textbf{A} = \{A_1(x), A_2(x),...,A_n(x)\}$, each is a function formed by the combinations of an activation function, not only ReLU as in ReLU-KAN. Given an input $x$, two trainable parameters: phase low $l_i$ and phase high $h_i$, the function $A_i(x)$ is the combination of $act(x-l_i)$ and $act(h_i-x)$ with $act$ is an activation function. For short, the result of $act(x-l_i)$ and $act(h_i-x)$ are denoted as $p$ and $q$ respectively. \Cref{tab:function_type} lists different function types which  $A_i(x)$ can be. We limit our work to functions of degree three or lower; for higher-order functions~\cite{so2024higher}, a memory error may occur in some problems, including image classification.

\begin{table*}[ht]
	\caption{Function combination types of $A_i(x)$ in AF-KAN, with the default type being \texttt{quad1}, applied in ReLU-KAN with ReLU activations.}
	\centering
	\begin{tabular}{p{3cm}p{5cm}p{3cm}}
            \hline
		\textbf{Function type} & \textbf{Formula} &  \textbf{Short name} \\
    \hline 
    linear & $p + q$ & \texttt{sum} \\
    bilinear & $p \times q$ & \texttt{prod} \\
    bilinear & $p + q + (p \times q$) & \texttt{sum\_prod} \\
    quadratic & $ {(p \times q)}^2 $ & \texttt{quad1} \\
    quadratic & $ {p^2 + q^2  + (p \times q)}^2 $ & \texttt{quad2} \\
    cubic & $ {(p + q)\times (p^2 + q^2)} $ & \texttt{cubic1} \\
    cubic & $ {(p \times q)}^3 $ & \texttt{cubic2} \\
            %\multicolumn{5}{l}{\texttt{attn} = attention mechanism, \texttt{conv} = convolutional layers, \texttt{conv\&pool} = convolutional \& pooling layers}  \\
             \hline
	\end{tabular}
	\label{tab:function_type}
\end{table*}

In ReLU-KAN, the function domain is assumed to be within the range $[0-1]$~\cite{qiu2024relu}, requiring input data to be normalized accordingly before being fed into the network. However, since we use activation functions beyond ReLU, the behavior of the function $\mathbf{A}$ may vary, leading to different function shapes, and the normalization constant $c$ is no longer appropriate. 
%Moreover, real-world datasets contain diverse input values. While normalizing these data often improves model performance, in some cases, retaining the original range may help preserve meaningful feature representations.

\begin{figure*}[htbp]
  \centering
\includegraphics[scale=0.6]{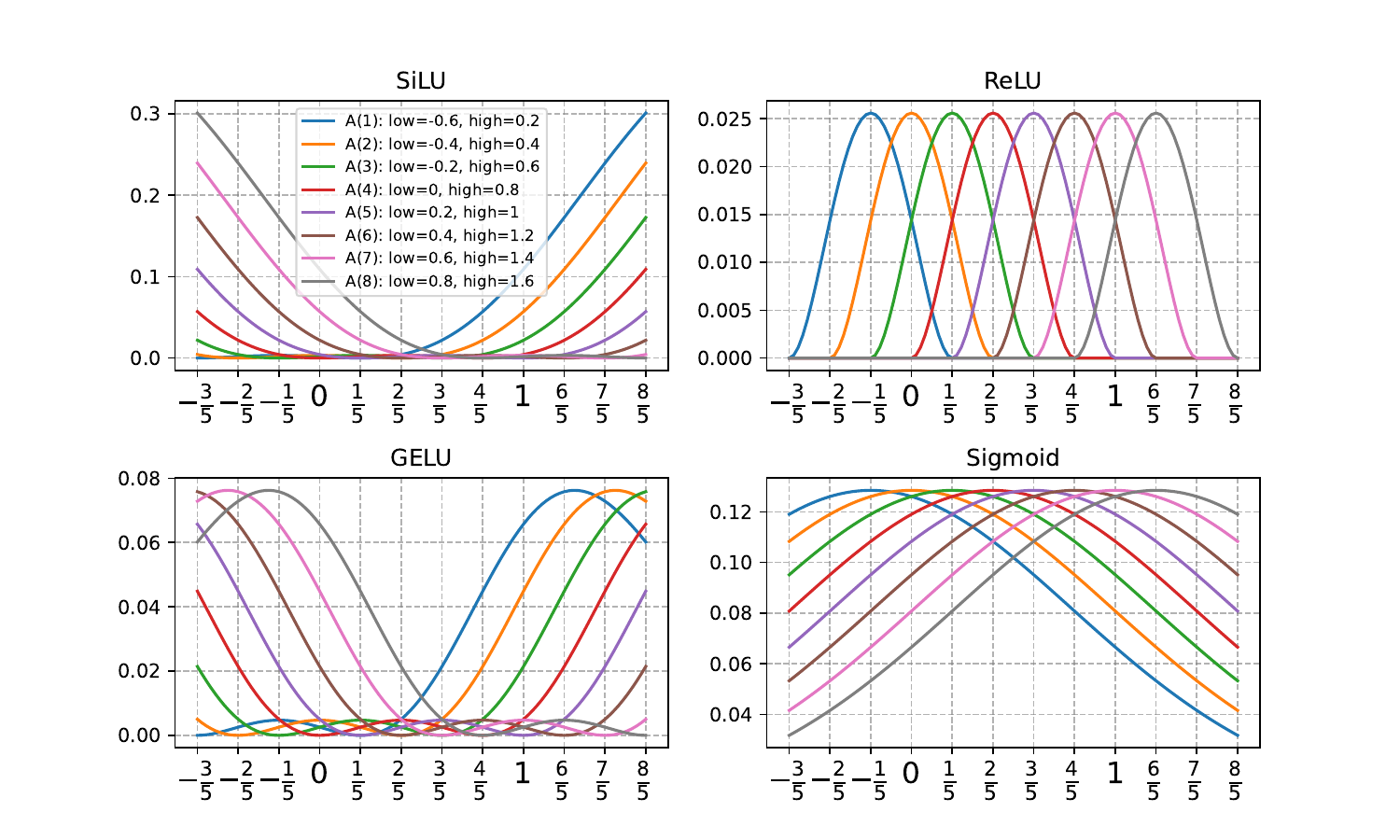}
  \centering
  \caption{Simulate the plots of the function $\textbf{A}$ using several activation functions. This function is configured with a grid size of $G = 5$, a spline order of $k = 3$, and the function type \texttt{quad1}.}
\label{fig:function_plots}
\end{figure*}

To demonstrate the various forms of function $\mathbf{A}$, we take a simple example with activation functions—SiLU, ReLU, GELU, and Sigmoid—using the \texttt{quad1} form, originally introduced in ReLU-KAN. This example replicates the values used in the simulation in \Cref{fig:relu_kan}, where the grid size is $G = 5$ and the spline order is $k = 3$. However, instead of using $x \in [0,1]$, we extend the range to $x \in [-3/5, 8/5]$ to illustrate a broader spectrum of function behavior. The low-phase and high-phase values are set to $[-3/5, -2/5, -1/5, 0, 1/5, 2/5, 3/5, 4/5]$ and $[1/5, 2/5, 3/5, 4/5, 1, 6/5, 7/5, 8/5]$, respectively. As shown in \Cref{fig:function_plots}, the plots for SiLU, GELU, and Sigmoid differ significantly from ReLU, making it challenging to identify a common maximum point as the normalization constant for all functions $A_i(x)$. From a computational perspective, we prefer a simple normalization method to handle different function types more effectively.

Instead of using the normalization constant $c$, we substitute it with L2 normalization and min-max scaling to normalize the data values. Given that  
$o = A_i(x)$ is the function output for an input data $x$, we can apply the following formulas to normalize its value:

\begin{subequations}
\begin{equation}
    \text{Normalize using L2 norm: } \\ 
    o = \frac{o}{\|o\|_2} = \frac{o}{\sqrt{\sum_{j=1}^{n} o_j^2}}, \text{with } n = |o|
    \label{eq:l2_norm} 
\end{equation}

\begin{equation}
\begin{aligned}
    \text{Min-max scaling, scale to [0,1]: } \\
    o = \frac{o - o_{\min}}{o_{\max} - o_{\min}}
    \label{eq:minmax_scaling}
\end{aligned}
\end{equation}
\end{subequations}
By default, the output value remains in the range \([0,1]\) throughout this normalization. However, we can set a different range for specific tasks as needed.

A study indicates that KANs use excessive parameters, resulting in underperformance compared to MLPs in many tasks when the same parameter count is used~\cite{yu2024kan}. Since ReLU-KAN inherits KAN with functions in the style of B-splines, its parameters do not change. However, in AF-KAN, we apply parameter reduction methods, as described in \Cref{sec:att_me}, to decrease the number of parameters to experiment with its ability with MLPs.

\subsection{Implementation of AF-KAN}

\begin{figure*}[htbp]
  \centering
\includegraphics[scale=0.7]{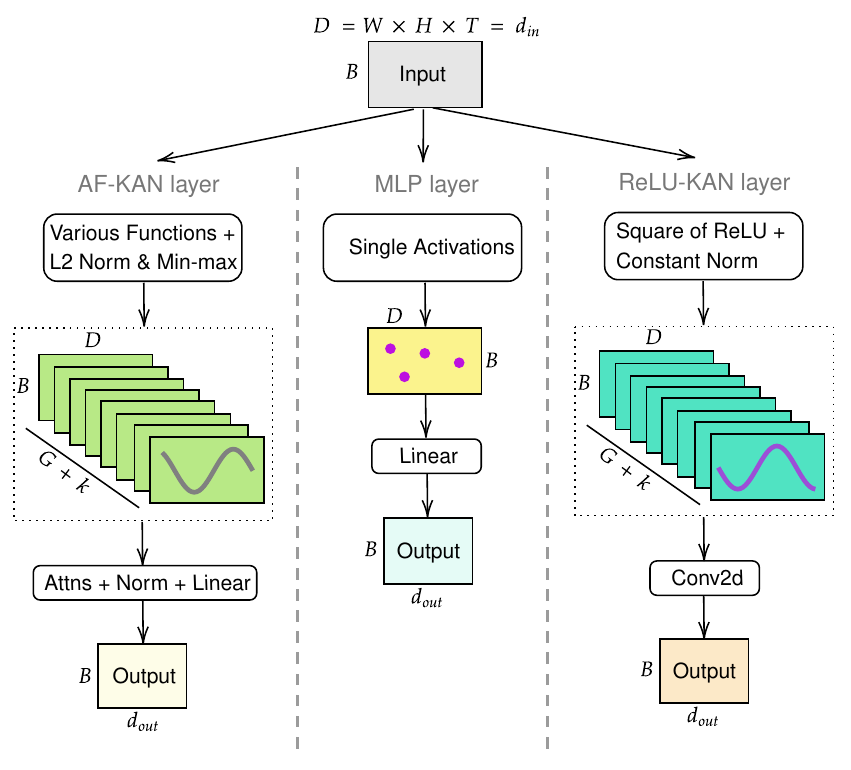}
  \centering
  \caption{Flow of an input through AF-KAN, MLP, and ReLU-KAN layers. \textbf{Left}: AF-KAN enables a broader range of functions using L2 norm, min-max scaling, and attention mechanisms to reduce parameters. \textbf{Center}: MLP applies a single activation function and linear transformation. \textbf{Right}: ReLU-KAN uses the "Square of ReLU" with a constant norm and a 2D convolutional layer.}
\label{fig:net_layers}
\end{figure*}

Different from the original ReLU-KAN~\cite{qiu2024relu}, which was designed for a single input, AF-KAN was developed to handle multiple inputs, making it suitable for various tasks, including image classification. AF-KAN mainly works well with single-channel data, \( T = 1 \). However, it can be modified to handle multi-channel data more effectively.

\Cref{fig:net_layers} illustrates how an input propagates through AF-KAN, MLP, and ReLU-KAN layers. In details for AF-KAN, given a set of inputs with (batch) size \( B \), each with a data dimension \( D \) (\( D = W \times H \times T \), where \( W \) is the width, \( H \) is the height, and \( T \) is the number of channels), the input tensor \( X \) has a shape of \( (B, D) \). Let \( d_{\text{in}} \) and \( d_{\text{out}} \) be the input and output dimensions of a layer, respectively. A strict condition here is that \( d_{\text{in}} \) is always equal to \( D \) in each layer to ensure a consistent dimensionality flow within the network. The output \( Y \) has a shape of \( (B, d_{\text{out}}) \) is generated when passing \( X \) within this layer.

In AF-KAN, the function  \( \textbf{A} \) uses phase low \( l \) and phase high \( h \) as compacted vectors:

\begin{subequations}
\begin{equation}
    \text{Phase low: } \\
     l = \left\{ \frac{-k}{G}, \frac{-k+1}{G}, \frac{-k+2}{G}, \dots, \frac{G-1}{G} \right\}
     \label{eq:phase_low} 
\end{equation}

\begin{equation}
\begin{aligned}
    \text{Phase high: } \\
    h &= l + (k + 1)/G \\ 
    &= \left\{ \frac{-k}{G} + \frac{k+1}{G}, \frac{-k+1}{G} + \frac{k+1}{G}, \frac{-k+2}{G} + \frac{k+1}{G}, \dots, \frac{G-1}{G} + \frac{k+1}{G} \right\} \\ &=
    \left\{ \frac{1}{G}, \frac{2}{G}, \frac{3}{G}, \dots, \frac{G+k}{G} \right\}
    \label{eq:phase_high} 
\end{aligned}
\end{equation}
\end{subequations}
which can expand to the size of the input \( X \) to perform element-wise operations on matrices in two terms \( act(X - l) \) and \( act(h - X) \), as well as their combinations. When passing through the function \( A \), the input \( X \) produces the output \( X_A \), which has the shape \( (B, D, G + k) \), where \( G \) is the grid size and \( k \) is the spline order.

Then, we normalize \( X_A \) using L2 norm and min-max scaling as in \Cref{eq:l2_norm} and \Cref{eq:minmax_scaling}. This step is performed in mini-batches. Although it affects each batch locally, it helps avoid the complexity of finding a global normalization constant for normalizing the input value range across various function types. After that, we will get the data tensor \( X_{N_1} \) with the same data dimensions as \( X_A \).

It is possible to apply a linear transformation to \( X_{N_1} \) with transformed shape \( (B, D \times (G + k)) \) by multiplying it with a weight matrix of shape \( (D \times (G + k), d_{\text{out}}) \) (and adding a bias matrix) to obtain the final output \( Y \) with shape \( (B, d_{\text{out}}) \). However, this way results in a weight matrix with too many parameters. To reduce parameters, we follow \citet{ta2025prkan} to \textbf{convert \( X_{N_1} \) from \( (B, D, G + k) \) to \( (B, D) \)} using attention before a linear transformation with a smaller weight matrix \( (D, d_{\text{out}}) \). Since normalization centers the data, we omit the bias matrix.

\subsection{Parameter reduction methods in AF-KAN}
\label{sec:att_me}
The ReLU-KAN implementation uses a 2D convolutional layer to extract information from the function output and transform it into a compatible format for subsequent layers\footnote{\url{https://github.com/quiqi/relu_kan}}. However, while this method preserves the number of parameters, it may not be the most effective for capturing essential features. 

Inspired by the experiments of \citet{ta2025prkan}, we improve feature extraction by applying two attention mechanisms to the function output: global attention and spatial attention. These mechanisms are selected for their ability to improve feature representation while maintaining computational efficiency during training. We exclude other attention mechanisms, such as local attention, self-attention, scaled dot-product attention, and multi-head attention~\cite{vaswani2017attention}, as they introduce higher computational costs and slower processing times, which can hinder model training on large datasets.

\subsubsection{Global Attention Mechanism}

As the name suggests, when computing the attention weight, global attention considers all elements in a given tensor. \Cref{eq:input_global_attn,eq:linear_transformation_global_attn,eq:softmax_global_attn,eq:elementwise_multiplication_global_attn,eq:summation_global_attn,eq:normalization_global_attn,eq:final_linear_global_attn} define how to apply this attention. From the output of the function \( \mathbf{A} \) after data normalization \( X_{N_1} \), its last dimension is reduced by performing a linear transformation to get \( X_{\text{linear}} \). Then, apply softmax on its data dimension \( D \) divided by a temperature scaling \( \tau \) to get the attention weight matrix \( W_{\text{attn}} \). This learnable parameter is used to control the sharpness or smoothness of the softmax output. To maintain numerical stability, the expression \(\max(\tau, 1.0)\) ensures that the value does not fall below 1.0, preventing potential instability in computations. We set the default value as the square root of the dimension size of \( D \).

%A higher value of \( \tau \) leads to a more uniform softmax distribution, ensuring that all values contribute more equally. Conversely, when \( \tau \) is lower, the distribution becomes more concentrated, allowing only a few values to dominate. 

Next, we obtain \( X' \) by multiplying \( X_{N_1} \) by the attention weight matrix \( W_{\text{attn}} \). Then, we reduce its last dimension to get \( X'' \) which has a shape of \( (B, D) \). We perform another data normalization (either batch normalization or layer normalization) on \( X'' \) to get \( X_{N2} \). This step ensures that the data values remain within a scale that is easy to train after some matrix operations. From here, the process is similar to an MLP layer. We perform a linear transformation by multiplying the output of \( X_{N2} \) passed through an activation function \( \sigma \) with a weight matrix \( W_{\text{out}} \) and/or adding a bias \( b_{\text{out}} \) to get the final output \( X_{\text{out}} \), which has a shape of \( (B, d_{\text{out}}) \). 

\begin{subequations}
\begin{equation}
    \text{Input tensor: } \\
    X_{N_1} \in \mathbb{R}^{B \times D \times (G + k)} \label{eq:input_global_attn}
\end{equation}

\begin{equation}
    \text{Linear transformation: } \\
    X_{\text{linear}} = W_{\text{gk}} \times X_{N_1} + b_{gk}, \quad X_{\text{linear}} \in \mathbb{R}^{B \times D \times 1} \label{eq:linear_transformation_global_attn}
\end{equation}

\begin{equation}
    \text{Softmax over the data dimension $D$: } \\
    W_{\text{attn}} = \textit{softmax} \left( \frac{X_{\text{linear}}}{\max(\tau, 1.0)}, \text{dim} = -2 \right), \tau =\sqrt{|D|}, \quad W_{\text{attn}} \in \mathbb{R}^{B \times D \times 1} \label{eq:softmax_global_attn}
\end{equation}

\begin{equation}
    \text{Element-wise multiplication: } \\
    X' = X_{N_1} \odot W_{\text{attn}}, \quad X' \in \mathbb{R}^{B \times D \times (G + k)} \label{eq:elementwise_multiplication_global_attn}
\end{equation}

\begin{equation}
    \text{Summation along the last dimension: } \\
    X'' = \sum_{\text{dim}=-1} X', \quad X'' \in \mathbb{R}^{B \times D} \label{eq:summation_global_attn}
\end{equation}

\begin{equation}
    \text{Data Normalization: } \\
    X_{N_2} = norm(X''), \quad X_{N_2} \in \mathbb{R}^{B \times D}
    \label{eq:normalization_global_attn}
\end{equation}

\begin{equation}
    \text{Final linear transformation: } \\
    X_{\text{out}} = W_{\text{out}} \times \sigma(X_{N_2}) + b_{\text{out}}, \quad X_{\text{out}} \in \mathbb{R}^{B \times d_{\text{out}}} \label{eq:final_linear_global_attn}
\end{equation}
\end{subequations}

\subsubsection{Spatial Attention Mechanism}

This attention exploits spatial data to extract the attention weight matrix, described in \Cref{eq:input_spatial_attn,eq:permute_spatial_attn,eq:conv1d_spatial_attn,eq:softmax_spatial_attn,eq:elementwise_multiplication_spatial_attn,eq:summation_spatial_attn,eq:normalization_spatial_attn,eq:final_linear_spatial_attn}. From an input \( X_{N_1} \) with shape \( (B, D, G + k) \), we consider the dimension \( G + k \) as spatial data. Then, we permute and pass it through a convolutional layer to get \( X_{\text{conv}} \). On this tensor, we perform softmax on the dimension \( G + k \) and divide it by a temperature scaling \( \tau \), similar to its use in global attention. 

Next, we perform an element-wise multiplication between \( X_{N_1} \) and the attention weight matrix \( W_{\text{attn}} \) after permutation to get \( X' \). We then sum the last dimension of \( X' \) and perform data normalization to get \( X_{N_2} \) with shape \( (B, D) \). From here, similar to global attention, we perform a linear transformation to get the final output \( X_{\text{out}} \) with shape \( (B, d_{\text{out}}) \).

\begin{subequations}
\begin{equation}
    \text{Input tensor: } \\
    X_{N_1} \in \mathbb{R}^{B \times D \times (G + k)} \label{eq:input_spatial_attn}
\end{equation}

\begin{equation}
    \text{Permute tensor: } \\
    X_{\text{perm}} = permute(X_{N_1}), \quad X_{\text{perm}} \in \mathbb{R}^{B \times (G + k) \times D} \label{eq:permute_spatial_attn}
\end{equation}

\begin{equation}
    \text{1D convolution: } \\
    X_{\text{conv}} = conv1d(X_{\text{perm}}), \quad X_{\text{conv}} \in \mathbb{R}^{B \times (G + k) \times D} \label{eq:conv1d_spatial_attn}
\end{equation}

\begin{equation}
    \text{Softmax over spatial dimension $D$: } \\
    W_{\text{attn}} = \textit{softmax} \left( \frac{X_{\text{conv}}}{\max(\tau, 1.0)}, \text{dim} = -2 \right), \tau =\sqrt{|D|}, \quad W_{\text{attn}} \in \mathbb{R}^{B \times (G + k) \times D} \label{eq:softmax_spatial_attn}
\end{equation}

\begin{equation}
    \text{Element-wise multiplication: } \\
    X' = X_{N_1} \odot permute(W_{\text{attn}}), \quad X' \in \mathbb{R}^{B \times D \times (G + k)} \label{eq:elementwise_multiplication_spatial_attn}
\end{equation}

\begin{equation}
    \text{Summation along the last dimension: } \\
    X'' = \sum_{\text{dim}=-1} X', \quad X'' \in \mathbb{R}^{B \times D} \label{eq:summation_spatial_attn}
\end{equation}

\begin{equation}
    \text{Data Normalization: } \\
    X_{N_2} = norm(X''), \quad X_{N_2} \in \mathbb{R}^{B \times D}
    \label{eq:normalization_spatial_attn}
\end{equation}

\begin{equation}
    \text{Final linear transformation: } \\
    X_{\text{out}} = W_{\text{out}} \times \sigma(X_{N_2}) + b_{\text{out}}, \quad X_{\text{out}} \in \mathbb{R}^{B \times d_{\text{out}}} \label{eq:final_linear_spatial_attn}
\end{equation}
\end{subequations}

\subsubsection{Multi-step Linear Transformation}

Besides applying attention mechanisms to reduce the number of parameters used in AF-KAN, we can also apply a multi-step linear transformation. As mentioned, whenever obtaining the normalized function output \( X_{N_1} \) with shape \( (B, D, G + k) \), we apply a linear transformation by multiplying it (in shape of \( (B, D \times (G + k)) \)) with a weight matrix of shape \( (D \times (G + k), d_{\text{out}}) \) to get the output shape \( (B, d_{\text{out}}) \). However, this weight matrix has too many parameters. 

Instead, we perform two steps to transform \( X_{N_1} \) with shape \( (B, D, G + k) \) to the output shape \( (D, d_{\text{in}}) \) described in \Cref{eq:input_multi,eq:linear_transform_multi,eq:reshape_multi,eq:normalization_multi,eq:final_linear_multi}. First, we perform a linear transformation to get \( X_{\text{linear}} \) by multiplying it with a weight matrix \( W_{gk} \) and a bias. Then, we reshape and apply data normalization on \( X_{\text{linear}} \) to get \( X_{N_2} \). Second, we apply another linear transformation to transform \( X_{N_2} \) to \( X_{\text{out}} \) with shape \( (B, d_{\text{out}}) \). In this way, we only need two smaller weight matrices \( (G + k, 1) \) and \( (D, d_{\text{out}}) \) with/without their biases, instead of one with the shape of \( (D \times (G + k), d_{out}) \). Although we perform two steps for linear transformation, we still refer to this method as "multi-step linear transformation" to generalize the case.

\begin{subequations}
\begin{equation}
    \text{Input tensor: } \\
    X_{N_1} \in \mathbb{R}^{B \times D \times (G + k)} \label{eq:input_multi}
\end{equation}

\begin{equation}
    \text{First Linear transformation: } \\
    X_{\text{linear}} = W_{\text{gk}} X_{N_1} + b_{\text{gk}}, \quad X_{\text{linear}} \in \mathbb{R}^{B \times D \times 1} \label{eq:linear_transform_multi}
\end{equation}

\begin{equation}
    \text{Reshape operation: } \\
    X_{\text{reshaped}} = reshape(X_{\text{linear}}, B, -1) \label{eq:reshape_multi}
\end{equation}

\begin{equation}
    \text{Data normalization: } \\
    X_{N_2} = norm(X_{\text{reshaped}}), \quad X_{N_2} \in \mathbb{R}^{B \times D} \label{eq:normalization_multi}
\end{equation}

\begin{equation}
    \text{Final linear transformation: } \\
    X_{\text{out}} = W_{\text{out}} \times \sigma(X_{N_2}) + b_{\text{out}}, \quad X_{\text{out}} \in \mathbb{R}^{B \times d_{\text{out}}} \label{eq:final_linear_multi}
\end{equation}
\end{subequations}

\section{Experiments}
\subsection{Datasets}
\begin{figure*}[htbp]
  \centering
\includegraphics[scale=0.6]{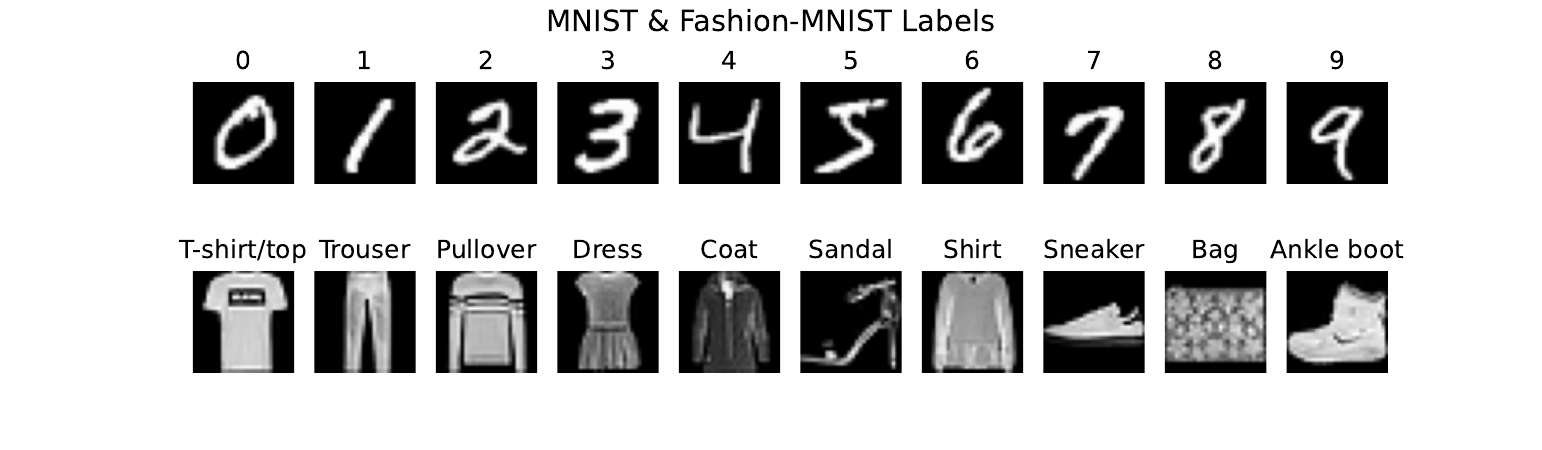}
  \centering
  \caption{The output labels of MNIST (first row) and Fashion-MNIST (second row).}
\label{fig:dataset_label}
\end{figure*}

We chose the MNIST and Fashion-MNIST datasets for the experiments due to their simplicity and structured format. Some of their samples by output labels are shown in \Cref{fig:dataset_label}. MNIST, derived from NIST, is a widely used resource for testing machine learning algorithms on handwritten digits~\cite{deng2012mnist}. It includes 60,000 training images and 10,000 test images, all size-normalized and centered in 28x28 pixel format. Each image is a binary vector of size 784. Given its simplicity, MNIST is ideal for quickly evaluating machine learning techniques and pattern recognition methods with minimal preprocessing.

The original MNIST dataset of handwritten digits is widely used for benchmarking image-based machine-learning methods. To offer more challenging and practical alternatives, Fashion-MNIST is a new dataset comprising 70,000 grayscale images, each measuring 28 × 28 pixels, representing fashion products across 10 categories, with 7,000 images per category~\cite{xiao2017fashion}. The dataset is divided into a training set of 60,000 images and a test set of 10,000 images. Fashion-MNIST is intended to serve as a direct replacement for the original MNIST dataset, providing a benchmark for machine learning algorithms with the same image size, data format, and structure for training and testing splits.

%Each dataset includes 28×28 grayscale images across 10 output labels, to accurately identify the label for each image. The datasets are divided into training and validation sets of 60,000 and 10,000 samples, respectively. MNIST features handwritten digits, while Fashion-MNIST contains images of clothing items, providing a balanced and challenging foundation for evaluating different image classification models.

\subsection{Training Configuration}
\label{sec:training_conf}

\begin{table*}[ht]
	\caption{The number of used parameters and FLOPs in models with approximately the same number of parameters and the same network structure. The network structures are applied to training on both MNIST and Fashion-MNIST.}
	\centering
	\begin{tabular}{p{1.5cm}p{4cm}p{3cm}p{2.5cm}p{2cm}}
            \hline
	\textbf{Group} &	\textbf{Model} & \textbf{Network structure} & \textbf{\#Used Params} & \textbf{\#FLOPs} \\
    \hline 
    \multirow{5}{1.5cm}{1}  
& AF-KAN-global\_attn & (784, 64, 10) & \textbf{52626} & \textbf{37.32K}  \\
& AF-KAN-spatial\_attn	& (784, 64, 10) & \textbf{52626} & \textbf{37.32K} \\
& AF-KAN-multistep	& (784, 64, 10) & \textbf{52626} & \textbf{37.32K} \\
& MLP & (784, 64, 10) & 52512 & \textbf{1.844K} \\
& PRKAN & (784, 64, 10) & 52604 & 20.36K \\
 
            \hline
            \multirow{8}{1.5cm}{2}  
& BSRBF-KAN & (784, 7, 10) & 51588 & 3.16K \\
& EfficientKAN & (784, 7, 10) & \textbf{55580} & \textbf{1.582K} \\
& FastKAN & (784, 7, 10) & \textbf{51605} & 12.74K \\
& FasterKAN & (784, 8, 10) & 52382 & 26.92K \\
& FC-KAN & (784, 8, 10) & 53968 & 3.16K \\
& ReLU-KAN & (784, 9, 10) & 52411 & \textbf{104.82K} \\
            \hline
            \multirow{6}{1.5cm}{3}  
& BSRBF-KAN & (784, 64, 10) & 459024 & 3.4K \\
& EfficientKAN & (784, 64, 10) & 508160 & \textbf{1.696K} \\
& FastKAN & (784, 64, 10) & 459098 & 103.48K \\
& FasterKAN & (784, 64, 10) & 408206 & 28.84K \\
& FC-KAN & (784, 64, 10) & \textbf{560656} & 3.4K \\
& ReLU-KAN & (784, 64, 10) & \textbf{315146} & \textbf{630.3K} \\

            \hline
            \multicolumn{4}{l}{\texttt{global\_attn} = global attention, \texttt{spatial\_attn} = spatial attention} \\
              \multicolumn{4}{l}{\texttt{multistep} = multistep linear transformation} \\
             \hline
	\end{tabular}
	\label{tab:model_params_flops}
\end{table*}

The MLP is regarded as the standard model, characterized by a network structure of (784, 64, 10) comprising 784 input neurons, 64 hidden neurons, and 10 output neurons, totaling 52,604 parameters. It utilizes SiLU and layer normalization as default settings. To ensure a fair comparison, we categorize the models into three groups, as presented in \Cref{tab:model_params_flops}. We use the \texttt{ptflops}\footnote{\url{https://pypi.org/project/ptflops/}} package to compute the number of MACs (multiply–accumulate operations) and then multiply that by 2 to determine the number of FLOPs.
\begin{itemize}
    \item \textbf{Group 1}: This group includes models that share the same network structure and a similar number of parameters, such as AF-KAN variants, PRKAN, and MLP. All models are designed with approximately 52k parameters. Because of differences in network structures, some models may have slightly higher or lower parameter counts. While AF-KANs and PRKAN have slightly higher parameter counts than MLPs, they require significantly more FLOPs—19 times and 11 times more, respectively. 
    \item \textbf{Group 2}: In this group, all models need to have their structures modified to achieve a similar number of parameters as MLPs. EfficientKAN has the highest parameter count but the lowest FLOPs, while FastKAN has the fewest parameters. ReLU-KAN, on the other hand, requires the most FLOPs.
    \item \textbf{Group 3}: Not only limited to approximately the same parameters, we also design models with the same structure as AF-KANs and MLPs but requiring 6 to 10 times more parameters to demonstrate the effectiveness of AF-KANs. The number of parameters in our models differs slightly from the work of \citet{ta2024fc} because we count only the used parameters rather than the total parameters. In this group, FC-KAN with the function combinations obtains the most parameters while having the second least number of FLOPs, following only EfficientKAN. Notably, ReLU-KAN has the fewest parameters in this group, but it requires an astonishing 631.98K FLOPs, primarily due to the use of 2D convolutional layers.
\end{itemize}

We apply a consistent set of hyperparameters in all experiments: \texttt{batch\_size=64}, \texttt{learning\_rate=1e-3}, \texttt{weight\_decay=1e-4}, \texttt{gamma=0.8}, \texttt{optimizer=AdamW}, and \texttt{loss=CrossEntropy}. For ReLU-KAN and AF-KAN, we use \texttt{grid\_size=3} and \texttt{spline\_order=3} due to their optimal performance with these values, detailed in \Cref{sec:grid_and_order}. In other models, we use \texttt{grid\_size=5}, \texttt{spline\_order=3}, and \texttt{num\_grids=8}.

For FC-KAN, we use a combination of B-splines and DoGs within a quadratic function representation. PR-KAN sets the attention mechanism as the default method to reduce the number of parameters, based on the global attention mechanism but differing slightly from AF-KAN. In AF-KAN, we set SiLU as the activation function, use the function type \texttt{quad1}, apply layer normalization, and incorporate the global attention mechanism as default settings.

The training consists of 25 epochs on MNIST and 35 epochs on Fashion-MNIST, balancing convergence with computational efficiency. Each model is trained over five independent runs, and we calculate the average values for metrics such as training accuracy, validation accuracy, F1 score, and training time to minimize variability. All experiments are conducted on an RTX 3060 Ti GPU with 8GB of VRAM. Finally, we reused some results from two other works~\cite{ta2025prkan, ta2024fc}, which were trained on the same device.

\subsection{Same Parameters}
\begin{table*}[ht]
	\caption{The comparison of AF-KANs versus PRKANs, MLPs, and other KANs with approximately 52k parameters.}
	\centering
	\begin{tabular}{p{1.5cm}p{3.2cm}p{1.5cm}p{1.8cm}p{1.8cm}p{1.8cm}p{1.8cm}}
            \hline
		\textbf{Dataset} &  \textbf{Model}  & \textbf{Norm.} & \textbf{Train. Acc.} & \textbf{Val. Acc.} & \textbf{F1} & \textbf{Time (sec)} \\
    \hline 
    \multirow{12}{1.5cm}{\textbf{MNIST}}
    & AF-KAN-global\_attn & layer & 99.80 ± 0.08 & \textbf{97.89 ± 0.04} & \textbf{97.86 ± 0.04} & 224.12 \\
    & AF-KAN-spatial\_attn & layer & 99.79 ± 0.01 & 97.50 ± 0.05 & 97.46 ± 0.05 & 223.84 \\
    & AF-KAN-multistep & layer & \textbf{99.89 ± 0.03} & 97.49 ± 0.09 & 97.45 ± 0.09 & 205.94 \\ % should recheck 205.94
    & PRKAN-attn~\cite{ta2025prkan} & batch & 98.97 ± 0.33 & 97.29 ± 0.10 & 97.25 ± 0.10 & 179.35 \\
    &  PRKAN-attn~\cite{ta2025prkan} &  layer &  99.81 ± 0.09 &  97.46 ± 0.06 & 97.42 ± 0.06 &  178.02 \\
    &  MLP~\cite{ta2025prkan} &  layer &  99.84 ± 0.04 &  97.72 ± 0.05 &  97.69 ± 0.05 &  162.58 \\
    & BSRBF-KAN~\cite{ta2025prkan} & layer & 95.33 ± 0.14 & 92.83 ± 0.19 & 92.68 ± 0.19 & 209.73 \\
    & EfficientKAN~\cite{ta2025prkan} & -- & 93.33 ± 0.05 & 92.35 ± 0.12 & 92.24 ± 0.12 & 180.79 \\
    & FastKAN~\cite{ta2025prkan} & layer & 95.00 ± 0.11 & 93.10 ± 0.22 & 92.97 ± 0.24 & 164.64 \\
    & FasterKAN~\cite{ta2025prkan} & layer & 92.82 ± 0.06 & 92.30 ± 0.08 & 92.17 ± 0.09 & \textbf{155.30} \\
    & FC-KAN & layer & 97.47 ± 0.32 & 95.07 ± 0.07 & 94.98 ± 0.07 & 244.78 \\
    %& ReLU-KAN & layer  & 95.70 ± 0.41 & 92.40 ± 0.09 & 92.26 ± 0.10 & 170.52 \\
    %& ReLU-KAN & batch & 84.09 ± 2.06 & 83.30 ± 0.93 & 82.75 ± 1.00 & 176.39  \\
    & ReLU-KAN & -- & 97.51 ± 0.88 & 92.49 ± 0.08 & 92.35 ± 0.08 & 174.97 \\
            \hline
            \hline
    \multirow{12}{1.5cm}{\textbf{Fashion-MNIST}} 
    & AF-KAN-global\_attn & layer & 93.91 ± 0.05 & \textbf{89.30 ± 0.06} & \textbf{89.23 ± 0.07} & 311.48 \\

    & AF-KAN-spatial\_attn & layer & 93.89 ± 0.06 & 89.26 ± 0.04 & 89.21 ± 0.04 & 312.27 \\
    & AF-KAN-multistep & layer & 93.80 ± 0.05 & 89.25 ± 0.07 & 89.17 ± 0.06 & 314.29 \\
    & PRKAN-attn~\cite{ta2025prkan} & batch & 94.10 ± 0.17 & 88.87 ± 0.06 & 88.82 ± 0.06 & 259.18 \\ 
    &  PRKAN-attn~\cite{ta2025prkan} &  layer &  93.30 ± 0.20 &  88.82 ± 0.09 &  88.75 ± 0.10 &  250.62 \\
    &  MLP~\cite{ta2025prkan} &  layer &  94.20 ± 0.09 & 88.96 ± 0.05 &  88.92 ± 0.05 &  226.79 \\
    & BSRBF-KAN~\cite{ta2025prkan} & layer & 92.89 ± 0.07 & 86.82 ± 0.08 & 86.77 ± 0.08 & 295.71 \\
    & EfficientKAN~\cite{ta2025prkan} & -- & 89.00 ± 0.08 & 86.16 ± 0.12 & 86.07 ± 0.12 & 254.25 \\
    & FastKAN~\cite{ta2025prkan} & layer & 91.59 ± 0.07 & 87.34 ± 0.05 & 87.28 ± 0.04 & 229.21 \\
    & FasterKAN~\cite{ta2025prkan} & layer & 89.16 ± 0.09 & 86.67 ± 0.12 & 86.57 ± 0.11 & \textbf{217.42} \\
    & FC-KAN & layer & \textbf{94.61 ± 0.13} & 88.01 ± 0.02 & 87.97 ± 0.01 & 342.82 \\
    %& ReLU-KAN & layer & 91.58 ± 1.12 & 85.76 ± 0.06 & 85.65 ± 0.08 & 243.99 \\
    %& ReLU-KAN & batch & 89.01 ± 0.61 & 84.58 ± 0.10 & 84.39 ± 0.08 & 243.81 \\
    & ReLU-KAN & -- & 93.62 ± 0.61 & 85.15 ± 0.14 & 84.99 ± 0.15 & 246.26 \\
            \hline
             \multicolumn{7}{l}{Norm. = Data Normalization, Train. Acc. = Training Accuracy, Val. Acc. = Validation Accuracy }  \\
             \multicolumn{7}{l}{\texttt{attn} = attention, \texttt{global\_attn} = global attention, \texttt{spatial\_attn} = spatial attention}  \\
              \multicolumn{7}{l}{\texttt{multistep} = multistep linear transformation}\\
             
             %\multicolumn{7}{p{15cm}}{We select the best model performance based on data normalization positions (1 or 2) as structured in \Cref{fig:prkan_data_norm} and experimented in \Cref{norm_positions}, \Cref{tab:norm_pos_mnist}, and \Cref{tab:norm_pos_fashion_mnist}.}  \\
             \hline
	\end{tabular}
	\label{tab:same_params}
\end{table*}

In the first experiment, as shown in \Cref{tab:same_params}, we use models from \textbf{Group 1} and \textbf{Group 2} for comparison. EfficientKAN and ReLU-KAN have no normalization due to their design, while BSRBF-KAN, FastKAN, and FasterKAN use layer normalization by default. PRKANs, FC-KANs, and MLPs are recommended to use layer normalization due to its benefits~\cite{ta2024fc,ta2025prkan}. 

First, we compare AF-KANs with MLPs and PRKANs, which belong to \textbf{Group 1}. Due to their simple design, MLPs require the shortest training time, followed by PRKANs and then AF-KANs. Although PRKANs incorporate attention mechanisms, their validation accuracy and F1 score are lower than those of MLPs. It is clear that AF-KANs achieve the highest validation accuracy and F1 score, outperforming both MLPs and PRKANs. However, they require the longest training time---approximately 37--38\% more than MLPs---which can be considered a reasonable trade-off between training time and model performance.

Next, we compare AF-KANs with \textbf{Group 2}, where they significantly outperform other KANs. Notably, FC-KAN, which also relies on function combinations, performed better than other KANs but still lagged considerably behind AF-KANs. Moreover, FC-KANs required even more training time than AF-KANs. We believe this is due to the use of unsupported GPU functions in FC-KANs, such as B-splines, RBFs, and wavelets, in contrast to the traditional activation functions used in AF-KANs. Furthermore, FC-KANs did not integrate attention mechanisms, which could have enhanced model performance.

In both \textbf{Group 1} and \textbf{Group 2}, AF-KANs generally outperform all other KANs, though they require more training time. All AF-KAN variants rank at the top in validation accuracy and F1 score on Fashion-MNIST. Notably, the AF-KAN variant with a global attention mechanism achieves the highest validation accuracy and F1 score, reaching 97.89\% and 97.86\% on MNIST, and 89.30\% and 89.23\% on Fashion-MNIST.  In another aspect, Faster-KAN has shown no notable performance improvements despite its faster training speed.

\subsection{Same Network Structure}
\begin{table*}[ht]
	\caption{The comparison of AF-KANs versus PRKANs, MLPs, and other KANs with the same network structure of (784, 64, 10).}
	\centering
	\begin{tabular}{p{1.3cm}p{3.2cm}p{1.5cm}p{2cm}p{1.8cm}p{1.8cm}p{1.8cm}}
            \hline
		\textbf{Dataset} &  \textbf{Model}  & \textbf{Norm.} & \textbf{Train. Acc.} & \textbf{Val. Acc.} & \textbf{F1} & \textbf{Time (sec)} \\
    \hline 
    \multirow{12}{1.5cm}{\textbf{MNIST}} 
    & AF-KAN-global\_attn & layer & 99.80 ± 0.08 & 97.89 ± 0.04 & 97.86 ± 0.04 & 224.12 \\

    & AF-KAN-spatial\_attn & layer & 99.79 ± 0.01 & 97.50 ± 0.05 & 97.46 ± 0.05 & 223.84 \\

    & AF-KAN-multistep & layer & 99.89 ± 0.03 & 97.49 ± 0.09 & 97.45 ± 0.09 & 205.94 \\ % should recheck 205.94
    
    & PRKAN-attn~\cite{ta2025prkan} & batch & 98.97 ± 0.33 & 97.29 ± 0.10 & 97.25 ± 0.10 & 179.35 \\
    
    &  PRKAN-attn~\cite{ta2025prkan} &  layer &  99.81 ± 0.09 &  97.46 ± 0.06 & 97.42 ± 0.06 &  178.02 \\
    
    &  MLP~\cite{ta2025prkan} &  layer &  99.84 ± 0.04 &  97.72 ± 0.05 &  97.69 ± 0.05 &  162.58 \\
    & BSRBF-KAN~\cite{ta2024fc} & layer & \textbf{100.00 ± 0.00} & 97.59 ± 0.02 & 97.56 ± 0.02 & 211.5 \\
    & EfficientKAN~\cite{ta2024fc}	& - & 99.40 ± 0.10 & 97.34 ± 0.05 & 97.30 ± 0.05 & 184.5 \\
    & FastKAN~\cite{ta2024fc}	& layer & 99.98 ± 0.01 & 97.47 ± 0.05 & 97.43 ± 0.05 & 164.47 \\
    & FasterKAN~\cite{ta2024fc}	 & layer & 98.72 ± 0.02 & 97.69 ± 0.04 & 97.66 ± 0.04 & \textbf{161.88} \\
    
    & FC-KAN~\cite{ta2024fc} & layer & \textbf{100.00 ± 0.00} & \textbf{97.91 ± 0.05} & \textbf{97.88 ± 0.05} & 263.29 \\
    %& ReLU-KAN & layer & \textbf{100.00 ± 0.00} & 96.31 ± 0.03 & 96.24 ± 0.03 & 179.99 \\
    %& ReLU-KAN & batch & - & - & - & - \\
    & ReLU-KAN & -- &  100.00 ± 0.00 & 96.74 ± 0.06 & 96.69 ± 0.06 & 174.06 \\
            \hline
            \hline
    \multirow{12}{1.5cm}{\textbf{Fashion-MNIST}} 
    & AF-KAN-global\_attn & layer & 93.91 ± 0.05 & 89.30 ± 0.06 & 89.23 ± 0.07 & 311.48 \\

    & AF-KAN-spatial\_attn & layer & 93.89 ± 0.06 & 89.26 ± 0.04 & 89.21 ± 0.04 & 312.27 \\

    & AF-KAN-multistep & layer & 93.80 ± 0.05 & 89.25 ± 0.07 & 89.17 ± 0.06 & 314.29 \\
    
    & PRKAN-attn~\cite{ta2025prkan} & batch & 94.10 ± 0.17 & 88.87 ± 0.06 & 88.82 ± 0.06 & 259.18 \\ 
    
    &  PRKAN-attn~\cite{ta2025prkan} &  layer &  93.30 ± 0.20 &  88.82 ± 0.09 &  88.75 ± 0.10 &  250.62 \\
    
    &  MLP~\cite{ta2025prkan} &  layer &  94.20 ± 0.09 & 88.96 ± 0.05 &  88.92 ± 0.05 &  226.79 \\

    & BSRBF-KAN~\cite{ta2024fc}	& layer & 99.34 ± 0.04 & 89.38 ± 0.06 & 89.36 ± 0.06 & 276.75  \\
    & EfficientKAN~\cite{ta2024fc}	& -- &	94.81 ± 0.09 & 88.98 ± 0.07 & 88.91 ± 0.08 & 247.85 \\
    & FastKAN~\cite{ta2024fc}	& layer &	98.25 ± 0.07 & 89.40 ± 0.08 & 89.35 ± 0.08 & \textbf{208.68} \\
    & FasterKAN~\cite{ta2024fc}	& layer &	94.41 ± 0.03 & 89.31 ± 0.03 & 89.25 ± 0.02 & 220.7 \\
    & FC-KAN~\cite{ta2024fc} & layer & \textbf{99.54 ± 0.13} & \textbf{89.99 ± 0.09} & \textbf{89.93 ± 0.08} & 369.2 \\
    %& ReLU-KAN & layer & \textbf{100.00 ± 0.00} & 87.83 ± 0.13 & 87.77 ± 0.13 & 250.75  \\
    %& ReLU-KAN & batch & - & - & - & - \\
    & ReLU-KAN & -- & 99.24 ± 0.60 & 86.65 ± 0.09 & 86.59 ± 0.08 & 244.67   \\
            \hline
             \multicolumn{7}{l}{Norm. = Data Normalization, Train. Acc. = Training Accuracy, Val. Acc. = Validation Accuracy }  \\
             \multicolumn{7}{l}{\texttt{attn} = attention, \texttt{global\_attn} = global attention, \texttt{spatial\_attn} = spatial attention}  \\
              \multicolumn{7}{l}{\texttt{multistep} = multistep linear transformation}\\
             
             %\multicolumn{7}{p{15cm}}{We select the best model performance based on data normalization positions (1 or 2) as structured in \Cref{fig:prkan_data_norm} and experimented in \Cref{norm_positions}, \Cref{tab:norm_pos_mnist}, and \Cref{tab:norm_pos_fashion_mnist}.}  \\
             \hline
	\end{tabular}
	\label{tab:same_network}
\end{table*}

In this experiment, we maintain a consistent network structure while evaluating AF-KANs against other KANs, which generate considerably more parameters. From an alternative perspective, this experiment highlights the superiority of AF-KANs in optimizing parameter efficiency while maintaining comparable performance. For reference, we also present outcomes for PRKANs and MLPs in \Cref{tab:same_params}. Furthermore, some results concerning PRKANs, MLP, BSRBF-KAN, FastKAN, FasterKAN, and EfficientKAN are sourced from prior studies~\cite{ta2024fc,ta2025prkan}.

While FC-KANs deliver the best performance, they entail a higher parameter count and the longest training duration. AF-KANs, however, show that with fewer parameters—between 6 to 10 times less—they still maintain impressive validation accuracy and F1 scores, closely rivaling FC-KAN on MNIST and surpassing other KAN variants in specific scenarios. Nonetheless, their chief limitation is an extended training time, which is about 38\% to 49\% longer compared to other KANs.Regarding convergence, BSRBF-KAN, FC-KAN, and ReLU-KAN exhibit effective convergence, achieving 100\% training accuracy on MNIST. However, on Fashion-MNIST, none of the models reach this milestone, with FC-KAN achieving the highest accuracy at 99.54\%. Besides, FastKAN and FasterKAN exhibit the best rapid training, although their performance does not significantly differ.

\subsection{Training time and FLOPs}

\begin{figure*}[htbp]
  \centering
\includegraphics[scale=0.4]{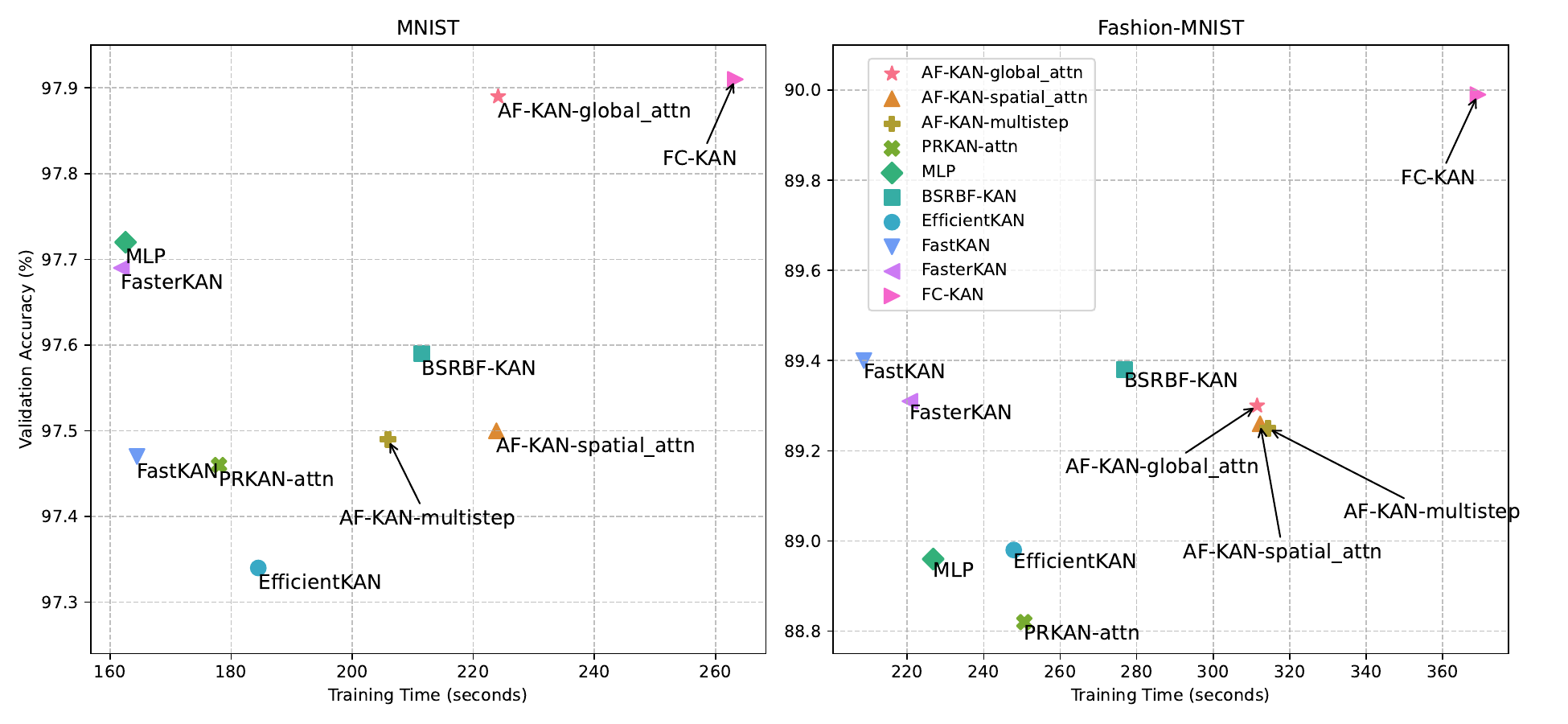}
  \centering
  \caption{The comparison of AF-KAN variants with other models of the same network structure in terms of training time and validation accuracy. AF-KANs, MLP, and PRKAN use approximately 52K parameters, while other models range from 400K to 560K parameters.}
\label{fig:acc_vs_time_same_structure}
\end{figure*}

\begin{figure*}[htbp]
  \centering
\includegraphics[scale=0.4]{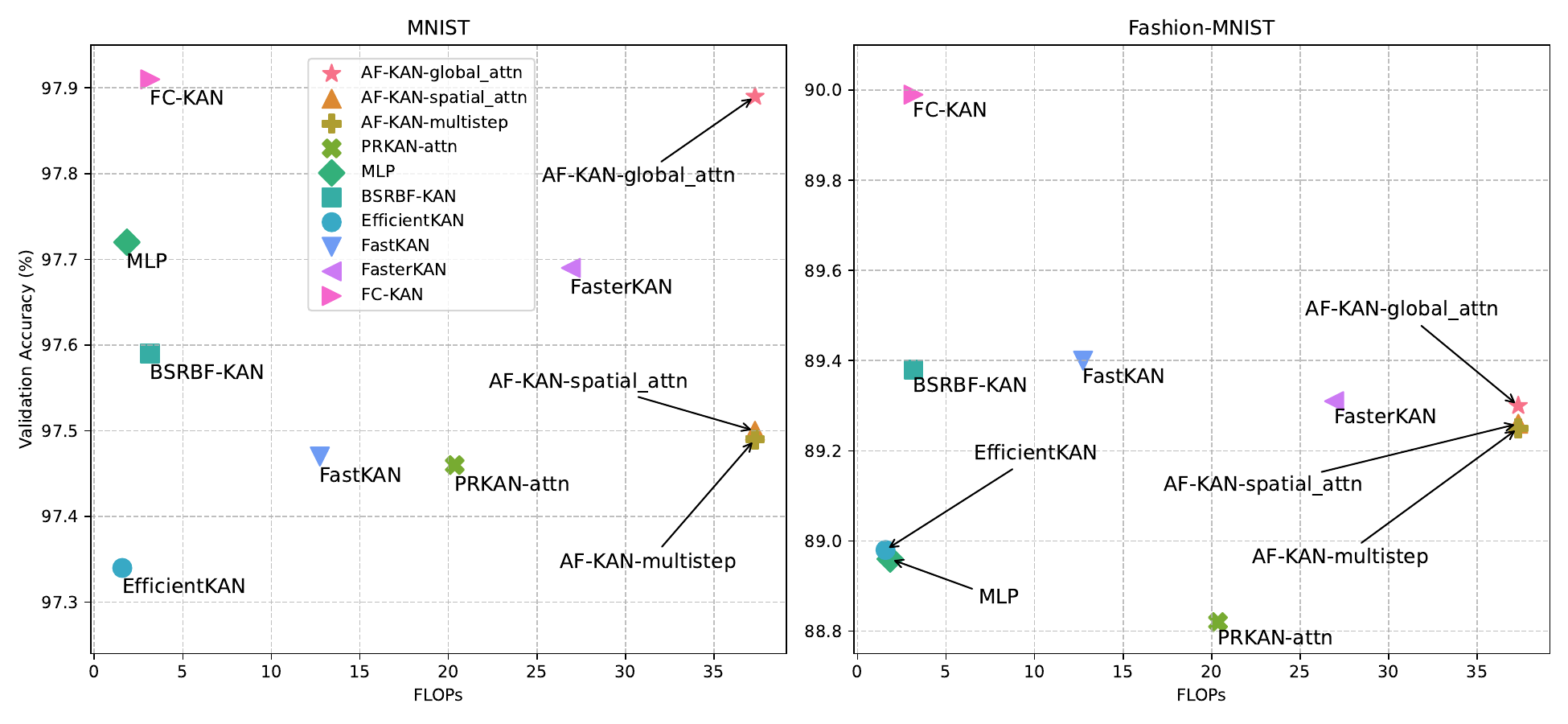}
  \centering
  \caption{The comparison of AF-KAN variants with other models of the same network structure in terms of flops and validation accuracy. AF-KANs, MLP, and PRKAN use approximately 52K parameters, while other models range from 400K to 560K parameters. }
\label{fig:acc_vs_flops_same_structure}
\end{figure*}

AF-KANs outperformed MLPs and other KANs with the same number of parameters, making this experiment less relevant. Therefore, we exclude it to focus on a more meaningful comparison. Instead, we compare AF-KANs with MLPs, PRKANs, and other KANs that share the same network structure. Only AF-KANs and PRKANs maintain parameter counts similar to MLPs, whereas other KANs introduce significantly more parameters despite their structural similarity.

\Cref{fig:acc_vs_time_same_structure} and \Cref{fig:acc_vs_flops_same_structure} illustrate the models' performance in two comparisons: validation accuracy versus training time and validation accuracy versus FLOPs. We exclude ReLU-KAN because it has the lowest accuracy while consuming the most FLOPs, making it the least effective model overall. In the first comparison, FC-KAN achieves the highest accuracy but also requires the longest training time, followed closely by AF-KANs on both datasets. On MNIST, AF-KAN-global-attn ranks second to FC-KAN in both training time and validation accuracy, while its variants position themselves in the middle, consistently outperforming some KANs. A similar trend is observed among AF-KAN variants on Fashion-MNIST. 

In the second comparison, FC-KAN achieves the highest accuracy while requiring the fewest FLOPs, whereas AF-KAN variants consume the most FLOPs. Despite this, they attain the second-highest accuracy on MNIST and remain moderately competitive with other models in terms of accuracy on Fashion-MNIST. Overall, AF-KANs demonstrate competitive performance compared to other KANs, even while using the same number of parameters as MLPs. However, optimizing AF-KANs for FLOP efficiency and reducing training time is essential for future improvements.

\subsection{Ablation Study}
In order to evaluate the influence of various AF-KAN components on model performance, we conduct multiple ablation studies focusing on activation functions, function types, grid size and spline order, as well as data normalization.

\subsubsection{Activation Functions}

\begin{table*}[ht]
	\caption{The comparison between AF-KAN variants by activation functions.}
	\centering
	\begin{tabular}{p{1.5cm}p{2.5cm}p{1.8cm}p{1.8cm}p{1.8cm}p{1.8cm}p{1.4cm}}
            \hline
		\textbf{Dataset} &  \textbf{Act. Func.}  & \textbf{Train. Acc.} & \textbf{Val. Acc.} & \textbf{F1} & \textbf{Time (sec)} \\
    \hline 
    \multirow{9}{1.5cm}{\textbf{MNIST}} 
    & ELU & 99.66 ± 0.14 & 97.81 ± 0.05 & 97.78 ± 0.05 & 224.4 \\
 & GELU & 99.89 ± 0.03 & 97.72 ± 0.05 & 97.69 ± 0.05 & 224.87 \\
 & Leaky RELU & \textbf{99.92 ± 0.03} & 97.30 ± 0.11 & 97.26 ± 0.11 & \textbf{220.7} \\
 & ReLU & 99.89 ± 0.02 & 97.21 ± 0.07 & 97.16 ± 0.08 & 224.46 \\
 & SELU & 99.85 ± 0.03 & 97.85 ± 0.02 & 97.82 ± 0.02 & 223.33 \\
 & Sigmoid & 96.78 ± 0.03 & 96.06 ± 0.08 & 96.00 ± 0.08 & 226.45 \\
 & SiLU & 99.80 ± 0.08 & \textbf{97.89 ± 0.04} & \textbf{97.86 ± 0.04} & 224.12 \\
 & Softplus & 98.59 ± 0.03 & 97.29 ± 0.04 & 97.25 ± 0.04 & 222.83 \\
 & Tanh & 99.62 ± 0.22 & 97.28 ± 0.08 & 97.25 ± 0.08 & 225.46 \\
            \hline
            \hline
    \multirow{9}{1.5cm}{\textbf{Fashion-MNIST}} 
     & ELU & 93.79 ± 0.05 & 89.21 ± 0.05 & 89.17 ± 0.06 & 313.13 \\
 & GELU & \textbf{94.38 ± 0.03} & \textbf{89.36 ± 0.08} & \textbf{89.29 ± 0.08} & 313.24 \\
 & Leaky RELU & 93.54 ± 0.09 & 89.00 ± 0.12 & 88.97 ± 0.12 & \textbf{309.83} \\
 & ReLU & 93.57 ± 0.06 & 88.95 ± 0.04 & 88.88 ± 0.03 & 315.94 \\
 & SELU & 93.42 ± 0.15 & 89.16 ± 0.03 & 89.08 ± 0.02 & 309.85 \\
 & Sigmoid & 88.84 ± 0.03 & 87.00 ± 0.02 & 86.86 ± 0.02 & 313.19 \\
 & SiLU & 93.91 ± 0.05 & 89.30 ± 0.06 & 89.23 ± 0.07 & 311.48 \\
 & Softplus & 89.64 ± 0.02 & 87.54 ± 0.06 & 87.42 ± 0.06 & 316.45 \\
 & Tanh & 93.89 ± 0.23 & 88.93 ± 0.08 & 88.85 ± 0.08 & 315.98 \\
            \hline
             \multicolumn{7}{l}{Act. Func. = Activation Function, Train. Acc = Training Accuracy, Val. Acc. = Validation Accuracy }  \\
             %\multicolumn{7}{l}{\texttt{attn} = attention mechanism}  \\
             \hline
	\end{tabular}
	\label{tab:act_funs}
\end{table*}

Since AF-KAN employs diverse activation functions to craft function types, choosing suitable ones not only speeds up training but also enhances model performance. In this experiment, we set up  AF-KANs with the function type \texttt{quad1}, including global attention mechanism and layer normalization. We then evaluate its performance using various activation functions, presented in \Cref{tab:act_funs}. Additional information about these functions can be found in \Cref{appendix:act_funcs}.

On the MNIST dataset, Leaky ReLU achieves the highest training accuracy and the fastest training time, while SiLU enables the model to attain the highest validation accuracy and F1 score. For the Fashion-MNIST dataset, GELU excels in both training and validation accuracy, whereas Leaky ReLU maintains the fastest convergence. Additionally, activation functions such as ELU and SELU exhibit competitive performance. In summary, this study supports the use of SiLU as a preferred choice, with ELU, GELU, and SELU also serving as effective alternatives. 
%The choosing of SiLU also be found in several works on KANs~\cite{liu2024kan,liu2024kan2.0}.

\subsubsection{Function Types}

Similar to the ablation study on activation functions, we set up AF-KANs with global attention mechanism and layer normalization while varying the function types. In this experiment, we compare different function types used in AF-KANs to justify our choice of \texttt{quad1} as the default, which is also the function type used in ReLU-KAN. Recall that \texttt{quad1} follows the form $(p \times q)^2$, where $p = act(x - l)$ and $q = act(h - x)$. For a full list of function types, refer to \Cref{tab:function_type}. AF-KANs offer flexibility in choosing function types, allowing experimenters to adapt them to their specific problems.

\Cref{tab:function_types} presents the performance of AF-KAN variants across different function types. Notably, function types do not significantly impact training speed, as their training times remain similar. However, it is evident that \texttt{quad1} achieves the best results on both datasets, followed by \texttt{sum} on MNIST and \texttt{sum} and \texttt{cubic1} on Fashion-MNIST. In summary, this study confirms the suitability of \texttt{quad1} as the default function type in AF-KANs.

\begin{table*}[ht]
	\caption{The comparison between AF-KANs variants by function types.}
	\centering
	\begin{tabular}{p{1.5cm}p{2.5cm}p{1.8cm}p{1.8cm}p{1.8cm}p{1.8cm}p{1.4cm}}
            \hline
		\textbf{Dataset} &  \textbf{Func. Type}  & \textbf{Train. Acc.} & \textbf{Val. Acc.} & \textbf{F1} & \textbf{Time (sec)} \\
    \hline 
    \multirow{7}{1.5cm}{\textbf{MNIST}} 
     & \texttt{quad1} & 99.80 ± 0.08 & \textbf{97.89 ± 0.04} & \textbf{97.86 ± 0.04} & 224.12 \\
     & \texttt{quad2} & 99.70 ± 0.06 & 97.70 ± 0.03 & 97.67 ± 0.03 & 229.4 \\
     & \texttt{sum} & 99.81 ± 0.02 & 97.88 ± 0.02 & 97.85 ± 0.02 & 222.07 \\
     & \texttt{prod} & \textbf{99.90 ± 0.05} & 97.64 ± 0.03 & 97.60 ± 0.03 & \textbf{219.55} \\
     & \texttt{sum\_prod} & 99.67 ± 0.04 & 97.72 ± 0.04 & 97.69 ± 0.04 & 225.49 \\
     & \texttt{cubic1} & 99.72 ± 0.09 & 97.74 ± 0.04 & 97.71 ± 0.04 & 225.83 \\
     & \texttt{cubic2} & 99.88 ± 0.03 & 97.56 ± 0.03 & 97.53 ± 0.04 & 223.22 \\
            \hline
            \hline
    \multirow{7}{1.5cm}{\textbf{Fashion-MNIST}} 
      & \texttt{quad1} & \textbf{93.91 ± 0.05} & \textbf{89.30 ± 0.06} & \textbf{89.23 ± 0.07} & 311.48 \\
      & \texttt{quad2} & 93.53 ± 0.04 & 88.90 ± 0.03 & 88.82 ± 0.04 & 319.49 \\
      & \texttt{sum} & 93.58 ± 0.07 & 89.06 ± 0.06 & 88.99 ± 0.06 & 311.01 \\
      & \texttt{prod} & 93.22 ± 0.12 & 88.69 ± 0.11 & 88.61 ± 0.11 & \textbf{309.28} \\
      & \texttt{sum\_prod} & 92.96 ± 0.10 & 88.88 ± 0.04 & 88.82 ± 0.04 & 317.63 \\
      & \texttt{cubic1} & 93.31 ± 0.06 & 89.06 ± 0.04 & 89.00 ± 0.05 & 319.74 \\
      & \texttt{cubic2} & 93.47 ± 0.05 & 88.97 ± 0.04 & 88.92 ± 0.04 & 312.63 \\
            \hline
             \multicolumn{7}{l}{Func. Type = Function Type, Train. Acc = Training Accuracy, Val. Acc. = Validation Accuracy }  \\
             %\multicolumn{7}{l}{\texttt{attn} = attention mechanism}  \\
             \hline
	\end{tabular}
	\label{tab:function_types}
\end{table*}

\subsubsection{Grid Size and Spline Order}
\label{sec:grid_and_order}
According to \citet{liu2024kan}, the spline order \( k \) of KANs adheres to a neural scaling law expressed as \( \alpha = k + 1 \). In cases where \( \alpha = 4 \) or \( k = 3 \), KANs reached saturation while exhibiting the most rapid scaling law in toy examples. Some works with KANs over image classification  choose a grid size of less than 10, usually 2, 3, and 5 \cite{li2024kolmogorov, wang2025efkan, chen2024lss, moradi2024kolmogorov, ta2025prkan}. The larger spline order or grid size increases the number of parameters used in a KAN and requires more training time. Therefore, it is crucial to select appropriate values for different problems.

\begin{figure*}[htbp]
  \centering
\includegraphics[scale=0.46]{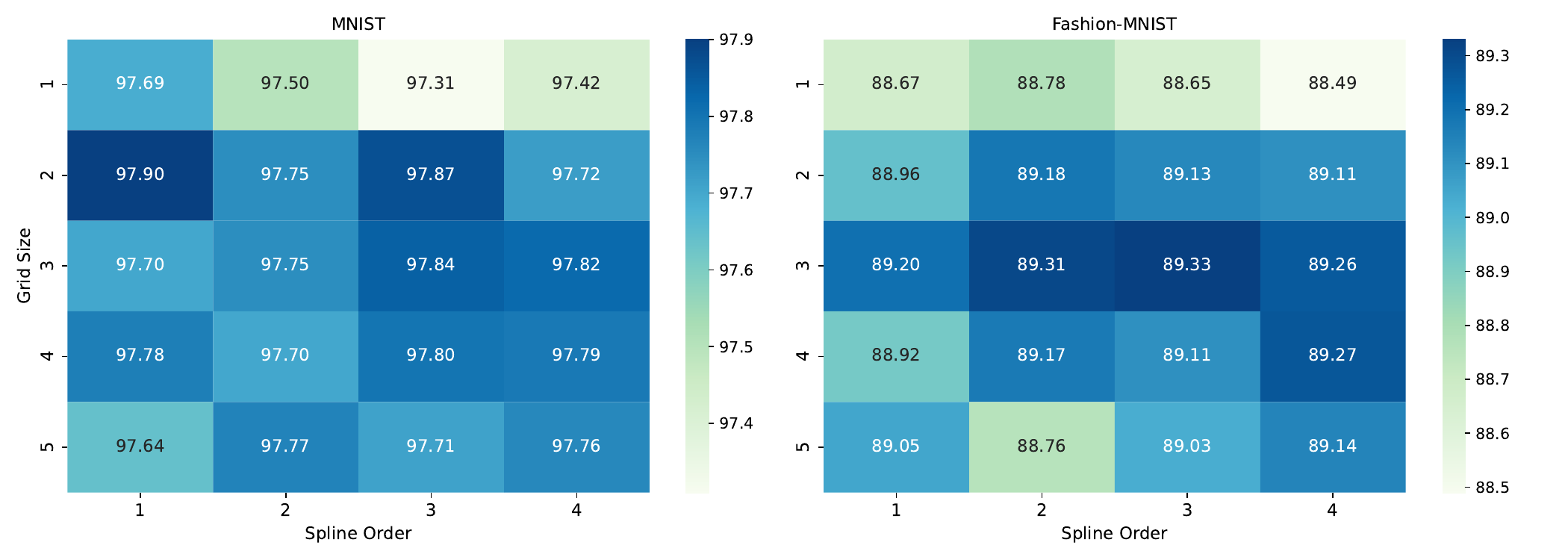}
  \centering
  \caption{The heatmap of AF-KAN validation accuracies for various grid sizes and spline orders.}
\label{fig:grid_order}
\end{figure*}

In this experiment, we compare the spline order \( k = \{1, 2, 3, 4\} \) and the grid size \( G = \{1, 2, 3, 4, 5\} \). Each model, defined by a specific grid size and spline order, is trained in 2 independent runs, and we take the average validation accuracy. The results are then presented as a heatmap in \Cref{fig:grid_order}. AF-KANs achieve the best accuracies with smaller grid sizes, preferably 2 or 3, and perform well with a spline order of 3. On Fashion-MNIST, the best result is obtained with a grid size of 1 and a spline order of 2. On MNIST, the best result is achieved with a grid size of 3 and a spline order of 3. In short, this experiment supports our use of grid size and spline order of 3 for the default settings of AF-KANs.

\subsubsection{Data Normalization}
\label{sec:data_norm}

\begin{table*}[ht]
	\caption{The impact of data normalization on AF-KANs.}
	\centering
	\begin{tabular}{p{1.5cm}p{1.5cm}p{1.5cm}p{1.8cm}p{1.8cm}p{1.8cm}p{1.8cm}p{1.4cm}}
            \hline
		\textbf{Dataset} &  \textbf{L2MM}  &  \textbf{PLN}  & \textbf{Train. Acc.} & \textbf{Val. Acc.} & \textbf{F1} & \textbf{Time (sec)} \\
    \hline 
    \multirow{6}{1.5cm}{\textbf{MNIST}} 
    & No & batch & 99.51 ± 0.11 & 97.43 ± 0.08 & 97.40 ± 0.08 & 211.03 \\
    & Yes & batch & \textbf{99.87 ± 0.03} & 97.60 ± 0.03 & 97.57 ± 0.03 & 228.55 \\
    & No & layer & 99.66 ± 0.09 & 97.79 ± 0.05 & 97.76 ± 0.05 & 208.93 \\
    & Yes & layer & 99.80 ± 0.08 & \textbf{97.89 ± 0.04} & \textbf{97.86 ± 0.04} & 224.12 \\
    & No & none & 91.03 ± 0.19 & 91.31 ± 0.16 & 91.15 ± 0.16 & \textbf{206.41} \\
    & Yes & none & 91.73 ± 0.15 & 91.82 ± 0.12 & 91.67 ± 0.12 & 220.66 \\
            \hline
            \hline
    \multirow{6}{1.5cm}{\textbf{Fashion-MNIST}} 
      & No & batch & 93.46 ± 0.04 & 88.81 ± 0.06 & 88.74 ± 0.06 & 294.38 \\
      & Yes & batch & \textbf{94.09 ± 0.08} & 89.12 ± 0.07 & 89.06 ± 0.08 & 319.52 \\
      & No & layer &  93.49 ± 0.08 & 89.21 ± 0.09 & 89.15 ± 0.09 & 289.97 \\
      & Yes & layer & 93.91 ± 0.05 & \textbf{89.30 ± 0.06} & \textbf{89.23 ± 0.07} & 311.48 \\
      & No & none & 84.89 ± 0.10 & 83.79 ± 0.12 & 83.67 ± 0.12 & \textbf{284.72} \\
      & Yes & none & 85.18 ± 0.07 & 83.89 ± 0.05 & 83.81 ± 0.06 & 309.06 \\
            \hline
             \multicolumn{7}{l}{L2MM = L2 norm and min-max scaling, PLN = Pre-linear normalization}  \\
             \multicolumn{7}{l}{Train. Acc = Training Accuracy, Val. Acc. = Validation Accuracy}  \\
             \hline
	\end{tabular}
	\label{tab:data_norm}
\end{table*}

In AF-KANs, normalization is applied in two places. First, we apply L2 normalization and the min-max scaling to normalize function outputs to the range $[0,1]$. Second, we apply pre-linear normalization, which is performed before the linear transformation in parameter reduction methods.  In this experiment, we use AF-KANs with the SiLU activation function, function type \texttt{quad1}, and a global attention mechanism to test the impact of these normalizations on the model performance.

As shown in \Cref{tab:data_norm}, data normalization significantly enhances model performance in terms of validation accuracy and F1-score, particularly when both L2MM (L2 norm and min-max scaling) and PLN (pre-linear normalization) are applied. The best AF-KAN model is achieved when L2MM and layer normalization in PLN are used. When neither normalization method is employed, all models perform the worst on both the MNIST and Fashion-MNIST datasets.

The exclusion of L2MM results in a minor performance degradation compared to its inclusion. In contrast, the absence of PLN leads to a significant decline in model performance.  This suggests that PLN has a more substantial impact on performance than L2MM. Furthermore, without L2MM, PLN shows competitive performance compared to the best model when layer normalization is used, but it experiences a decline with batch normalization. This suggests we can ignore L2MM but retain PLN with layer normalization if we need to save training time.

\section{Discussion}

ReLU-KAN is designed to enhance GPU parallelization~\cite{qiu2024relu}. In our experiments, it outperforms B-spline-based models like EfficientKAN in speed but remains slower than RBF- and RSWAF-based models such as FastKAN and FasterKAN. This slow training is likely due to using a 2D convolution layer or the expansion of inputs to facilitate matrix operations. Since ReLU cannot handle negative values, AF-KAN introduces a broader range of activation functions and function types to improve feature extraction. Moreover, AF-KAN enhances parameter efficiency even further by utilizing attention mechanisms, resulting in a parameter count similar to that of MLPs, as discussed in ~\citet{ta2025prkan}. Reducing parameters represents an emerging trend in KAN that showcases its true potential compared to MLPs.

Although AF-KANs show positive results, our work has several limitations:

\begin{itemize}
    \item \textbf{Simple datasets and shallow network structure}: The experiments use simple datasets (MNIST and Fashion-MNIST) with a shallow network structure (786, 64, 10). Since AF-KANs show effectiveness compared to other models, we expect similar trends in deeper networks and multi-channel datasets. However, the scalability of AF-KANs to complex datasets like CIFAR-10, CIFAR-100, or ImageNet remains an open question. Further investigations are needed to assess their feasibility in large-scale architectures, including transformer-based models or hybrid approaches incorporating convolutional layers.

    \item \textbf{Lack of comparison with fully equipped MLPs}: Our study primarily enhances AF-KANs while applying minimal modifications to MLPs, equipping them only with layer normalization. We also doubt that AF-KANs can compete with MLPs integrated with attention mechanisms regarding training time and other evaluation metrics. A more thorough comparison with MLPs enhanced by various architectural improvements could provide deeper insights into the advantages and limitations of AF-KANs.

    \item \textbf{Increased training time and FLOPs in AF-KANs}: While chasing model performance improvement, we must accept trade-offs in terms of training time and FLOPs. However, further optimization, such as pruning redundant function components or using low-rank approximations, could help balance performance and computational cost. Besides, exploring more hardware-efficient implementations, such as tensor decomposition methods or kernel fusion techniques, could improve AF-KAN.

    \item \textbf{Robustness and generalization}: While AF-KANs perform well in image classification tasks, their robustness in more challenging domains (e.g., adversarial settings, real-world noisy datasets) remains unclear. Investigating their generalization ability across diverse data modalities, such as time-series or tabular data, could further demonstrate their flexibility.
\end{itemize}

To further clarify the effectiveness of AF-KANs, we must test them on more complex network structures, a wider variety of datasets, and integrate similar components from AF-KANs into MLPs. The structure of AF-KAN also needs to be revised and evaluated to minimize training time and FLOPs while maintaining performance that is comparable to MLPs. Furthermore, given the function-based transformations in AF-KANs, exploring GPU-optimized implementations or specialized hardware adaptations could unlock their full potential for large-scale applications.

\section{Conclusion}
We introduced AF-KAN, a novel KAN developed based on ReLU-KAN, incorporating additional activation functions, function types, and data normalization. It applies parameter reduction methods, primarily attention mechanisms, which facilitates a parameter count comparable to MLPs. In our experiments, we conducted comparative analyses between models with approximately equivalent parameter counts and network structures. Following this, we performed a series of ablation studies on AF-KAN, examining activation functions, function types, grid sizes, spline orders, and data normalization methods to identify the optimal configuration.

AF-KAN significantly outperformed MLPs and other KANs in terms of approximate parameter count and remained competitive with other KANs possessing similar network structures, despite utilizing much fewer parameters. Furthermore, our observations indicate that AF-KAN performs optimally with SiLU and its quadratic function combination. Smaller grid sizes and third-order splines yielded the best results for AF-KAN. Data normalization techniques, such as layer normalization, were also found to be crucial in enhancing AF-KAN's performance. However, AF-KAN does present certain disadvantages, including longer training times and increased FLOPs. Adding extra components to enhance AF-KAN accounts for the longer training time and increased use of FLOPs, which is seen as a reasonable tradeoff.

Future work will focus on optimizing function combinations and architectural improvements to enhance KANs while keeping parameter counts comparable to MLPs. Furthermore, evaluating KANs across various domains, including image super-resolution, natural language processing, and scientific computing, will help assess their generalization and practical utility. We aim to establish KANs as a more efficient and scalable alternative for machine learning applications by addressing these aspects.

\section*{Acknowledgments}
We also acknowledge the support of (1) the Foundation for Science and Technology Development of Dalat University and (2) FPT University, Danang for funding this research.

\bibliographystyle{unsrtnat}
\bibliography{references}  %%% Uncomment this line and comment out the ``thebibliography'' section below to use the external .bib file (using bibtex) .

\newpage
\appendix
\section{Activation Functions}
\label{appendix:act_funcs}

\textbf{ELU (Exponential Linear Unit)}~\cite{clevert2015fast}: Minimizes bias shift, ensures a seamless transition, and allows customization through  $\alpha$.
\begin{equation}
    f(x) = 
    \begin{cases} 
    x & \text{if } x > 0 \\
    \alpha(e^x - 1) & \text{if } x \leq 0
    \end{cases}
    \label{eq:elu_equation} 
\end{equation}

\textbf{GELU (Gaussian Error Linear Unit)}~\cite{hendrycks2016gaussian}: Integrates smoothness with non-linearity, follows a probabilistic approach, and performs exceptionally well in NLP tasks.
\begin{equation}
    f(x) = 0.5x \left(1 + \tanh\left(\sqrt{\frac{2}{\pi}}(x + 0.044715x^3)\right)\right)
    \label{eq:gelu_equation}
\end{equation}

\textbf{Leaky ReLU (Leaky Rectified Linear Unit)}~\cite{xu2015empirical}: Enables gradients for \( x < 0 \), prevents dead neurons, and offers customization through \( \alpha \).
\begin{equation}
    f(x) = 
    \begin{cases} 
    x & \text{if } x > 0 \\
    \alpha x & \text{if } x \leq 0
    \end{cases}
    \label{eq:leaky_relu_equation}
\end{equation}

\textbf{ReLU (Rectified Linear Unit)}~\cite{xu2015empirical}: Efficient with sparse activations but carries the risk of dead neurons.
\begin{equation}
    f(x) = \max(0, x)
    \label{eq:relu_equation}
\end{equation}

\textbf{SELU (Scaled Exponential Linear Unit)}~\cite{klambauer2017self}: Maintains self-normalization, produces scaled output, and depends on specific initialization.
\begin{equation}
    f(x) = \lambda 
    \begin{cases} 
    x & \text{if } x > 0 \\
    \alpha(e^x - 1) & \text{if } x \leq 0
    \end{cases}
    \label{eq:selu_equation}
\end{equation}

\textbf{Sigmoid}~\cite{han1995influence}: Smooth and bounded, maps input to \( (0,1) \), and is widely used in binary classification.
\begin{equation}
    f(x) = \frac{1}{1 + e^{-x}}
    \label{eq:sigmoid_equation}
\end{equation}

\textbf{SiLU (Sigmoid Linear Unit)}~\cite{elfwing2018sigmoid}: Smooth and self-gating, facilitating improved gradient flow.
\begin{equation}
    f(x) = \frac{x}{1 + e^{-x}}
    \label{eq:silu_equation}
\end{equation}

\textbf{Softplus (Smooth ReLU Approximation)}~\cite{ramachandran2017searching}: A smooth, differentiable alternative to ReLU that avoids a hard zero threshold.  

\begin{equation}
    f(x) = \log(1 + e^x)
    \label{eq:softplus_equation}
\end{equation}

%%% Uncomment this section and comment out the \bibliography{references} line above to use inline references.
% \begin{thebibliography}{1}

% 	\bibitem{kour2014real}
% 	George Kour and Raid Saabne.
% 	\newblock Real-time segmentation of on-line handwritten arabic script.
% 	\newblock In {\em Frontiers in Handwriting Recognition (ICFHR), 2014 14th
% 			International Conference on}, pages 417--422. IEEE, 2014.

% 	\bibitem{kour2014fast}
% 	George Kour and Raid Saabne.
% 	\newblock Fast classification of handwritten on-line arabic characters.
% 	\newblock In {\em Soft Computing and Pattern Recognition (SoCPaR), 2014 6th
% 			International Conference of}, pages 312--318. IEEE, 2014.

% 	\bibitem{hadash2018estimate}
% 	Guy Hadash, Einat Kermany, Boaz Carmeli, Ofer Lavi, George Kour, and Alon
% 	Jacovi.
% 	\newblock Estimate and replace: A novel approach to integrating deep neural
% 	networks with existing applications.
% 	\newblock {\em arXiv preprint arXiv:1804.09028}, 2018.

% \end{thebibliography}

\end{document}